\documentclass{article}

\IfFileExists{iclr2027_conference.sty}{%
  \usepackage{iclr2027_conference,times}%
  \IfFileExists{math_commands.tex}{%%%%% NEW MATH DEFINITIONS %%%%%

\usepackage{amsmath,amsfonts,bm}

\def\eqref#1{equation~\ref{#1}}
\def\1{\bm{1}}

\DeclareMathAlphabet{\mathsfit}{\encodingdefault}{\sfdefault}{m}{sl}
\SetMathAlphabet{\mathsfit}{bold}{\encodingdefault}{\sfdefault}{bx}{n}

}{}%
  \iclrfinalcopy  % arXiv build: print the real author block
}{%
  \usepackage[margin=1in]{geometry}%
  \usepackage{natbib}%
  \bibliographystyle{plainnat}%
  \typeout{^^JNOTE: iclr2027_conference.sty not found -- using article layout.^^J}%
}

\usepackage{amsmath,amssymb}
\usepackage{graphicx}
\graphicspath{{./}{figures/}}
\usepackage{booktabs}
\usepackage{multirow}
\usepackage{subcaption}
\usepackage{xcolor}
\usepackage[utf8]{inputenc}
\usepackage{hyperref}
\usepackage{url}

\usepackage{CJKutf8}
\newcommand{\zh}[1]{\begin{CJK}{UTF8}{gbsn}#1\end{CJK}}
\newenvironment{zhblock}{\begin{CJK}{UTF8}{gbsn}}{\end{CJK}}

\newcommand{\ShRange}{14--28\%}
\newcommand{\AffRange}{94--99\%}
\newcommand{\MaxOwners}{10}
\newcommand{\SaveRange}{22--44\%}
\newcommand{\ExpBaseK}{5}
\newcommand{\ExpAmpRange}{1.6--8.8$\times$}

\newcommand{\ExpTkTotal}{32}

\newcommand{\RuShChain}{6\%$\to$9\%$\to$14\%}

\newcommand{\RcShChain}{13\%$\to$20\%$\to$23\%$\to$24\%}

\newcommand{\MergeTotal}{1,102}
\newcommand{\MergeRecov}{147}
\newcommand{\MergeCov}{13\%}
\newcommand{\MergeChars}{6}
\newcommand{\MergeCharsMax}{116}
\newcommand{\MergeLossN}{44}
\newcommand{\MergeLossPct}{30\%}
\newcommand{\MergePrincN}{18}
\newcommand{\MergePrincPct}{12\%}
\newcommand{\MergeLossRange}{22--37\%}
\newcommand{\AffRcLow}{4}
\newcommand{\AffRcHigh}{19}
\newcommand{\LenRcLow}{23}
\newcommand{\LenRcHigh}{32}
\newcommand{\StoreRowsMax}{7,025}
\newcommand{\SpreadGoal}{0.03}

\newcommand{\SpreadNarr}{0.42}
\newcommand{\HeadGoal}{1\%}
\newcommand{\HeadNarr}{19\%}
\newcommand{\GoalLo}{0.96}
\newcommand{\GoalHi}{0.99}
\newcommand{\QualityLeads}{4}
\newcommand{\QualityCells}{16}
\newcommand{\AblOnEntries}{6,590}
\newcommand{\AblOffEntries}{20,669}
\newcommand{\AblFactor}{3.1}
\newcommand{\AblOffAff}{94\%}
\newcommand{\AblOnAff}{93\%}
\newcommand{\GoallessLo}{0.5\%}
\newcommand{\GoallessHi}{1.6\%}

\title{Agentsensus: Consensus-Compressed Shared Memory\\for Multi-Agent Story Worlds}

\author{Yu Pan\\
University of Nebraska--Lincoln\\
\texttt{yu.pan@unl.edu}}

\newcommand{\code}[1]{\texttt{\small #1}}

\begin{document}
\maketitle
% the style's finalcopy head claims publication; this is a preprint
\lhead{Preprint. Under review.}

\begin{abstract}
A agentic story world is a dynamic system simulating who learned what, when, and
from whom---yet the standard design gives each character a private memory
stream. A shared event is therefore stored once per witness, large duplication will be incurred in terms of storage.
We present Agentsensus, a story-world simulation framework in which there is an unified long-term memory. Records of the same event merge into one owned by all its
witnesses, and semantically relevant memory records are linked.
We evaluate on four worlds---two classical Chinese novels, \emph{Hamlet}, and
a real-world conflict timeline---run for 40 to 80 rounds against three
per-character memory designs under an equal-granularity protocol. Agentsensus
writes \SaveRange{} fewer entries than the closest baseline and is the only
design whose memory becomes shared (\ShRange{} of records held by more than
one character, some by \MaxOwners{}) and linked (\AffRange{}), at judged
simulation quality indistinguishable or even better than the baselines. An ablation attributes this to the
merge itself: disabling it multiplies the store by \AblFactor{}$\times$ and
takes sharing to exactly zero. Sharing also compounds with the horizon rather
than saturating early, rising \RuShChain{} as one world is re-run at 10, 20
and 40 rounds.
\end{abstract}

\section{Introduction}

A story-world simulation places dozens of LLM-driven characters in a world with
places, objects and documents, and lets a narrative emerge from what they say
and do to one another \citep{park2023generative,ran2025bookworld}. Unlike a
task-solving multi-agent system, where the team is judged on an answer it
produces \citep{wu2023autogen,hong2024metagpt,qian2024chatdev}, such a world is
judged on what it \emph{accumulates}: who learned what, when, and from whom.
Memory is therefore not a component of these systems but their substrate. It is
also the component that scales worst: a novel-seeded world starts with thousands
of canonical events already deposited, adds several hundred more per hundred
rounds, and must serve retrieval to every character at every round.

The dominant design, established by Generative Agents
\citep{park2023generative} and inherited by most subsequent frameworks
\citep{ran2025bookworld,zhang2024memsurvey}, gives each character a private,
append-only stream scored by recency, importance and relevance. This is faithful
to individual cognition, but as the substrate of a \emph{shared} world it
creates three structural problems, each worsening as the world grows.

\paragraph{P1: witness-multiplied redundancy.} Every shared experience is stored
once per participant: a war council of ten officers produces ten near-identical
records, and a novel-seeded world multiplies this by thousands of canonical
events. Storage, embedding cost and index size scale with the number of
\emph{witnesses} rather than of \emph{events}.

\paragraph{P2: fragmented, divergence-prone world knowledge.} The same event
exists as $N$ independently-worded copies that drift further apart as agents
paraphrase and reflect. No mechanism reconciles them, no record knows its
counterparts exist, and nothing connects a courier's report to the battle it
describes. The connective tissue of the story exists nowhere in the memory
substrate, so retrieval returns one agent's partial view even when the
collective holds the full picture.

\paragraph{P3: memory management that agents never perform.} Stream designs
expose maintenance operations---linking, revising, forgetting---and rely on
agents to use them. Empirically they do not: across two models and all four
backends we test, agents issued \emph{zero} calls to every discretionary
memory-management action, even with documentation, worked examples and an
id-free interface (\S\ref{sec:nomgmt}).

\paragraph{Our approach.} Agentsensus answers all three with one architectural
commitment: \textbf{the world's memory is a single store, and sharing is
computed, not assumed}. Memories stay owner-scoped---an agent recalls only what
it owns, so a shared store is not a shared mind---but when two agents record the
same event, an equivalence mechanism folds the records into one entry owned by
both. This is deduplication with a semantic rather than lexical criterion,
closer to record linkage \citep{fellegi1969record} and semantic corpus
deduplication \citep{abbas2023semdedup} than to caching, and it addresses P1 and
the divergence half of P2. Pieces split from one compound deposit are
automatically cross-linked, and recall expands one hop along those links into
what the caller also owns, giving the store the connective tissue P2 demands
without an agent ever building a graph. Because of P3, \emph{every} mechanism
runs inside \code{remember} and \code{recall} themselves.

\paragraph{Contributions.}
\begin{itemize}
\item A story-world simulation framework with a deterministic round-barrier
  scheduler, passive environments and information carriers, kernel-held
  conversation threads with distance-delayed delivery, and full-system
  checkpoints enabling bit-for-bit resumption.
\item A \textbf{consensus-compressed shared memory}: self-contained atomization,
  semantic-prefilter plus LLM-judged equivalence merging with owner-set union,
  automatic affiliation of split pieces, and auto-expanding owner-scoped recall.
\item An \textbf{equal-granularity, simulation-only evaluation protocol} against
  three faithful baseline reimplementations, applied to four worlds spanning two
  languages, three genres and both human and institutional actors.
\item A three-layer \textbf{case-study methodology} quantifying how the emergent
  memory structure tracks the story's social structure.

\end{itemize}

\section{Related Work}

\paragraph{Memory for a single agent.} Generative Agents
\citep{park2023generative} established the stream scored by recency, importance
and relevance and compressed into reflections. Variants change what is stored or
how it is maintained rather than who owns it: MemGPT \citep{packer2023memgpt}
pages context like an operating system, MemoryBank \citep{zhong2024memorybank}
decays it on a forgetting curve, Reflexion \citep{shinn2023reflexion} persists
verbal self-feedback, and A-MEM \citep{xu2025amem} links an agent's notes into an
evolving network. A survey of the area \citep{zhang2024memsurvey} finds the same
shape throughout: the memory belongs to one agent. Agentsensus keeps this
machinery but makes \emph{ownership} a property of a record rather than of a
store, which lets one record belong to several agents at once.

\paragraph{Multi-agent systems and shared state.} AutoGen
\citep{wu2023autogen}, CAMEL \citep{li2023camel}, MetaGPT
\citep{hong2024metagpt} and ChatDev \citep{qian2024chatdev} organize LLMs into
teams whose shared state is the conversation. That is adequate when a team works
a task to completion; a story world runs for hundreds of rounds, and no
transcript survives as a retrieval substrate at that length.

\paragraph{Memory shared across agents.} G-Memory \citep{zhang2025gmemory}
organizes a system's history into an insight/query/interaction hierarchy;
Collaborative Memory \citep{rezazadeh2025collaborative} shares fragments under
dynamic access control. Both are our baselines. Both keep every contributor's
record intact and mediate \emph{access} to them; neither merges records that say
the same thing across contributors. Consensus compression changes the number of
records rather than their visibility, and the owner set it produces is an access
policy that is derived rather than configured.

\paragraph{Deduplication and grounding.} Record linkage
\citep{fellegi1969record} formalized deciding whether two records denote one
entity; semantic deduplication \citep{abbas2023semdedup} showed embedding-space
near-duplicates can be removed without loss. What our setting adds is that the
duplicate carries a \emph{witness}: merging is also a statement that two
characters share a belief, which is why the merged record keeps a set of owners.
The psycholinguistic notion of grounding \citep{clark1991grounding} is the
phenomenon our owner sets approximate mechanically.

\paragraph{Retrieval over graphs.} RAG \citep{lewis2020rag} and its
graph-structured variants---HippoRAG \citep{gutierrez2024hipporag}, GraphRAG
\citep{edge2024graphrag}---build structure over a \emph{static} corpus in an
indexing pass with global knowledge of the documents. Our graph is built
incrementally at deposit time by agents who cannot see the store.

\paragraph{Story worlds and character agents.} Story generation has largely been
a planning problem \citep{mirowski2023dramatron,yang2022re3,yang2023doc};
character agents fit a model to one persona \citep{shao2023characterllm}.
Between them sit simulated worlds \citep{park2023generative,park2024thousand,
hua2023waragent}, and BookWorld \citep{ran2025bookworld}, which builds an agent
society from a novel---the same motivation as our sedimentation stage.
Agentsensus differs in what it puts under test: not the story that comes out,
but the memory substrate it comes from.

\section{The Agentsensus framework}
\label{sec:framework}

\begin{figure}[t]
\centering
\includegraphics[width=0.82\textwidth]{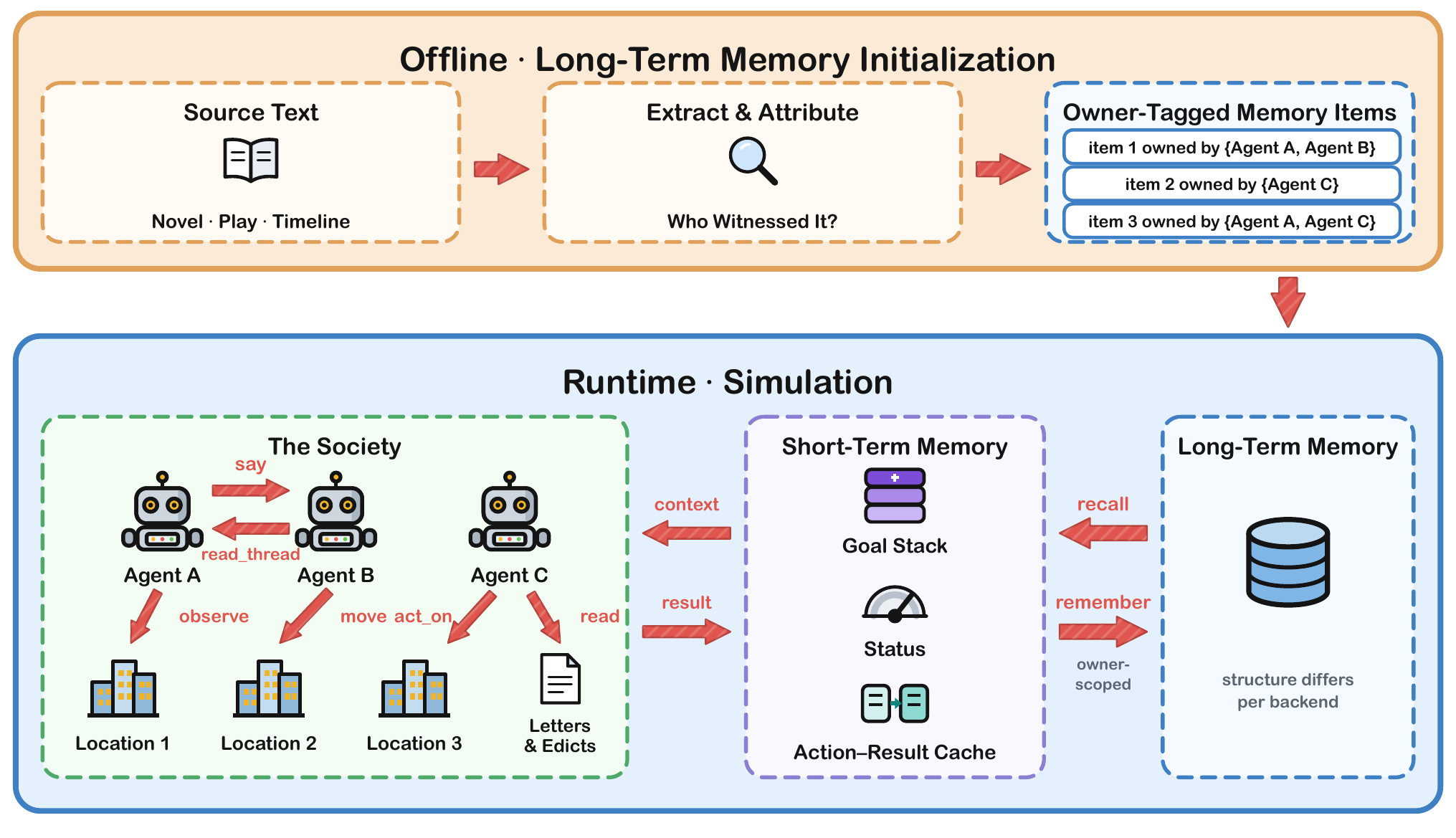}
\caption{The framework shared by all four backends. A source text is sedimented
into owner-tagged memory items that seed one long-term memory store; a round-barrier
kernel then steps every awake agent, and the actions they take deposit into and
retrieve from that store through short-term memory --- a goal stack, a status
register, and a cache of recent action--result pairs. The society panel names the six
actions that act on the world and on other agents. Only the long-term memory box differs between the
four systems compared in this paper.}
\label{fig:framework}
\end{figure}

\subsection{World model and round-barrier kernel}
Agents are of three kinds: \emph{characters} act, \emph{environments} (places)
own memories but never act, and \emph{information carriers} (letters, edicts,
scripts) hold text a character can read. A round is a barrier --- every awake
agent decides against the same world state, and effects become visible only in
the next round --- so a round's outcome does not depend on the order agents
were stepped in. Appendix~\ref{app:impl} gives the kernel, the action
repertoire and the short-term memory in full; none of it differs between the
four backends.

\subsection{Sedimentation}
Before simulation, the source text is turned into a world: a cast, a map,
carriers, and a memory store seeded with what those characters would already
know, each memory tagged with the agents that witnessed it. This is what makes
the redundancy of P1 measurable---the same canonical event arrives with several
owners.

\subsection{The consensus shared memory}
\label{sec:consensus}
A deposit \code{remember(agent, text)} is atomized into self-contained
statements capped at a token budget. Each statement is embedded and matched
against the whole store; candidates above a cosine threshold ($0.70$, top $5$)
go to an LLM judge, which either names an equivalent record or declines. On a
match, the record's owner set becomes the union of the two, its affiliated set
the union, and the shorter of the two texts is kept; on no match, a new record
is written. Statements atomized from one deposit are mutually affiliated, which
is where the memory graph comes from. Recall is owner-scoped and expands one hop
along affiliation into records the caller also owns.

\subsection{How the four designs differ}
\begin{figure}[t]
\centering
\includegraphics[width=0.78\textwidth]{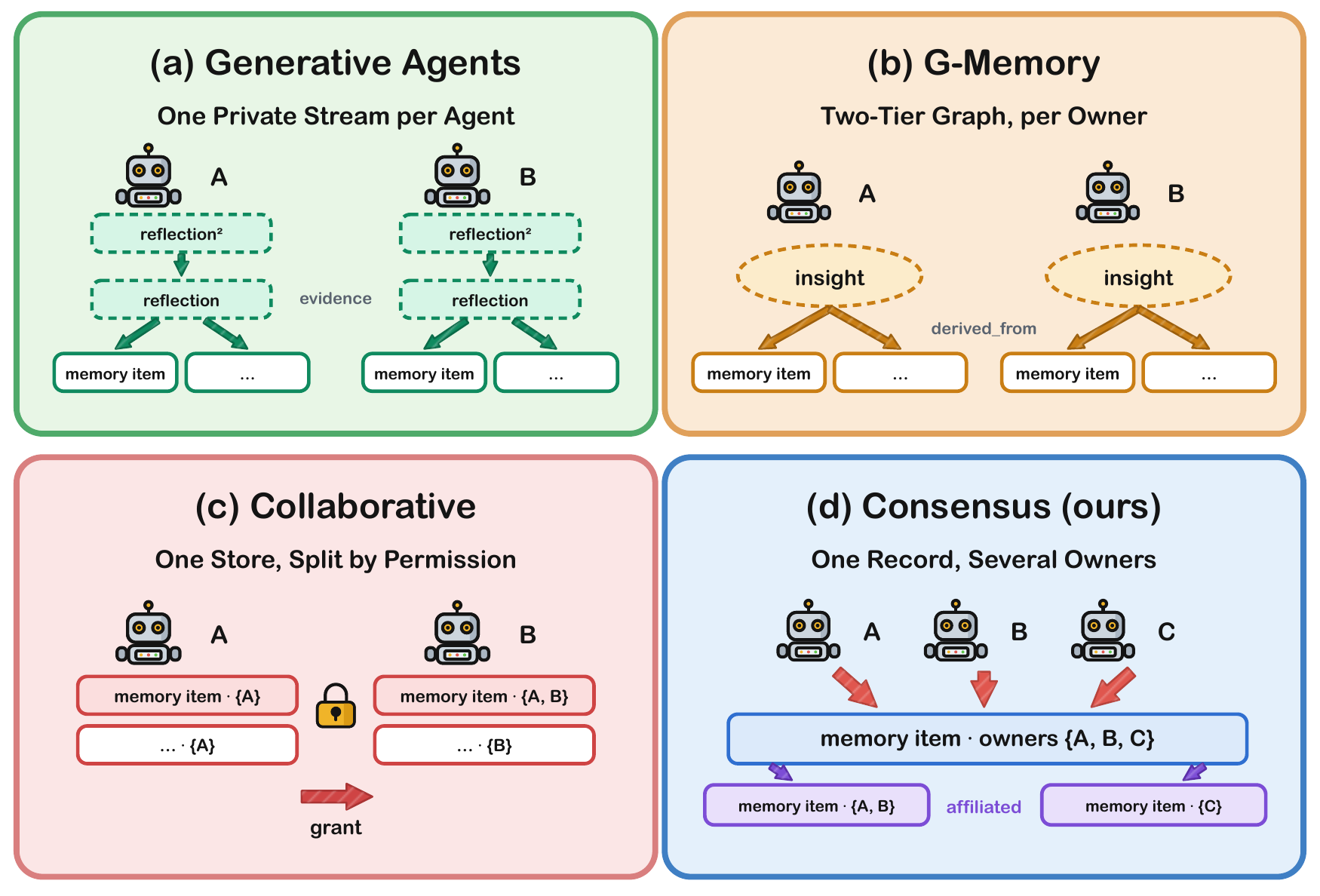}
\caption{The long-term memory structure of the four backends: private streams
scored by recency, importance and relevance, over which reflections are
distilled and themselves become evidence for further reflections; a two-tier
graph whose insights are distilled from the interaction rows they point back
to; a single store partitioned by access control, where sharing grants a second
reader rather than merging the rows; and the consensus store, where equivalent
deposits merge into one record whose owner set is the union of its witnesses,
with affiliation edges joining statements split from the same deposit.}
\label{fig:backends}
\end{figure}
Generative-Agents keeps a private stream per agent with importance scores and a
reflection tree. G-Memory keeps interaction and insight tiers joined by
\code{derived\_from} edges and retrieves at both levels. Collaborative keeps one
store partitioned by an access-control list. Only consensus merges records
across agents.

\section{Experimental setup}
\label{sec:setup}

\paragraph{Worlds.} Three Kingdoms (\emph{sanguo yanyi}) and Dream of the Red
Chamber (\emph{honglou meng}), sedimented from chapters 1--40 and run for 80
rounds; \emph{Hamlet}, sedimented from Acts I--III and run for 40; and a
Russia--Ukraine timeline through 2024-04, run for 40, whose agents are
institutions rather than people. The remainder of each work is held out as the
reference a continuation is scored against.

\paragraph{Baselines.} Faithful reimplementations of Generative-Agents-style
streams, G-Memory and Collaborative Memory, all sharing our kernel, personas,
scenarios and vector store, so the memory mechanism is the only difference.

\paragraph{Equal granularity.} Every backend routes \code{remember} through the
\emph{same} atomizer with the same self-containment requirement, so entry counts
are directly comparable and the only remaining difference is what each mechanism
does with identical atoms. All counts are simulation-only: the seeded history is
excluded, since it is identical by construction. It equalises what is
\emph{written}, not what is read: consensus recall expands one hop along
affiliation edges and no baseline does, so a consensus agent sees a median of
\ExpTkTotal{} rows per recall against \ExpBaseK{} for a baseline
(Appendix~\ref{app:cost}). The asymmetry favours consensus, which makes the
quality wash of \S\ref{sec:quality} the harder result to explain away.

\paragraph{Metrics.} Structural quantities (entries, multi-owner fraction,
affiliated fraction, merge depth) are deterministic. Continuation quality is
LLM-judged, three scorings each: \emph{grounding}, the fraction of the run's own
events consistent with canon; \emph{trajectory}, agreement of ten principals'
arcs with the held-out span; \emph{narrative}, a 1--5 rubric; and \emph{goal
pursuit}, the fraction of an agent's actions serving a goal it had itself pushed.
The first three read a rendered screenplay of the run---one language, an order of
magnitude smaller than the raw log---and the fourth reads the event log, since
goal management never becomes a dramatizable beat.

\section{Results}

\subsection{Footprint and structure}
\label{sec:structure}

\begin{table}[t]
\centering
\caption{Store size by backend and world. ``Sediment'' is the store at round~0, produced by each mechanism's own ingest of the identical source events --- one row per event under consensus, one row per witness under every baseline, plus each backend's own by-products (Generative-Agents reflections, G-Memory insight nodes). ``Sim'' is what the simulation then added and is the comparison the fairness protocol licenses: all four backends atomize identically, so entry counts are comparable by construction. Structure --- what fraction of the sim entries becomes shared and linked --- is Table~\ref{tab:structuremain}.}
\label{tab:footprint}
% generated
\begin{tabular}{lrrr}
\toprule
Backend & Sediment & Sim & Total \\
\midrule
\multicolumn{4}{l}{\emph{Three Kingdoms --- fiction, 80 rounds, 6,052 source events}} \\
\quad Consensus & \textbf{6,051} & \textbf{974} & 7,025 \\
\quad Gen.\ Agents & 20,896 & 1,251 & 22,147 \\
\quad G-Memory & 27,156 & 1,434 & 28,590 \\
\quad Collaborative & 19,856 & 1,455 & 21,311 \\
\midrule
\multicolumn{4}{l}{\emph{Red Chamber --- fiction, 80 rounds, 6,506 source events}} \\
\quad Consensus & \textbf{6,506} & \textbf{438} & 6,944 \\
\quad Gen.\ Agents & 27,306 & 785 & 28,091 \\
\quad G-Memory & 37,124 & 882 & 38,006 \\
\quad Collaborative & 26,681 & 789 & 27,470 \\
\midrule
\multicolumn{4}{l}{\emph{Russia--Ukraine --- real world, 40 rounds, 1,533 source events}} \\
\quad Consensus & \textbf{1,533} & \textbf{814} & 2,347 \\
\quad Gen.\ Agents & 9,492 & 1,268 & 10,760 \\
\quad G-Memory & 14,346 & 1,191 & 15,537 \\
\quad Collaborative & 8,493 & 1,091 & 9,584 \\
\midrule
\multicolumn{4}{l}{\emph{Hamlet --- fiction, 40 rounds, 1,135 source events}} \\
\quad Consensus & \textbf{1,135} & \textbf{130} & 1,265 \\
\quad Gen.\ Agents & 5,819 & 200 & 6,019 \\
\quad G-Memory & 7,068 & 193 & 7,261 \\
\quad Collaborative & 5,310 & 214 & 5,524 \\
\bottomrule
\end{tabular}

\end{table}

\begin{table}[t]
\centering
\caption{Sharing and linking across the four worlds, consensus store. Every
baseline is at exactly 0\% on both columns in every world --- by construction,
since no baseline merges records across agents or links atoms --- so they are
stated here rather than tabulated.}
\label{tab:structuremain}
% generated
\begin{tabular}{lrrr}
\toprule
World & Sim entries & Shared (multi-owner) & Linked (affiliated) \\
\midrule
Three Kingdoms (80r) & 974 & 185 (19\%) & 945 (97\%) \\
Red Chamber (80r) & 438 & 105 (24\%) & 412 (94\%) \\
Russia--Ukraine (40r) & 814 & 114 (14\%) & 806 (99\%) \\
Hamlet (40r) & 130 & 36 (28\%) & 127 (98\%) \\
\bottomrule
\end{tabular}

\end{table}

At equal granularity consensus writes \SaveRange{} fewer entries than the
closest baseline in each world (Table~\ref{tab:footprint}), because merging
folds witnesses together rather than storing them separately. It is also the
only backend whose memories become shared (\ShRange{} multi-owner) or linked
(\AffRange{}); the baselines sit at zero on both, by construction rather than by
tuning. The deepest merge reaches \MaxOwners{} witnesses---a presidential
air-defense directive co-owned by the president, his chief of staff and adviser,
the interior minister, the air-force and intelligence commanders, and the
security service.

\subsection{Growth and operation latency}
\label{sec:growth}

\begin{figure}[t]
\centering
\includegraphics[width=0.86\textwidth]{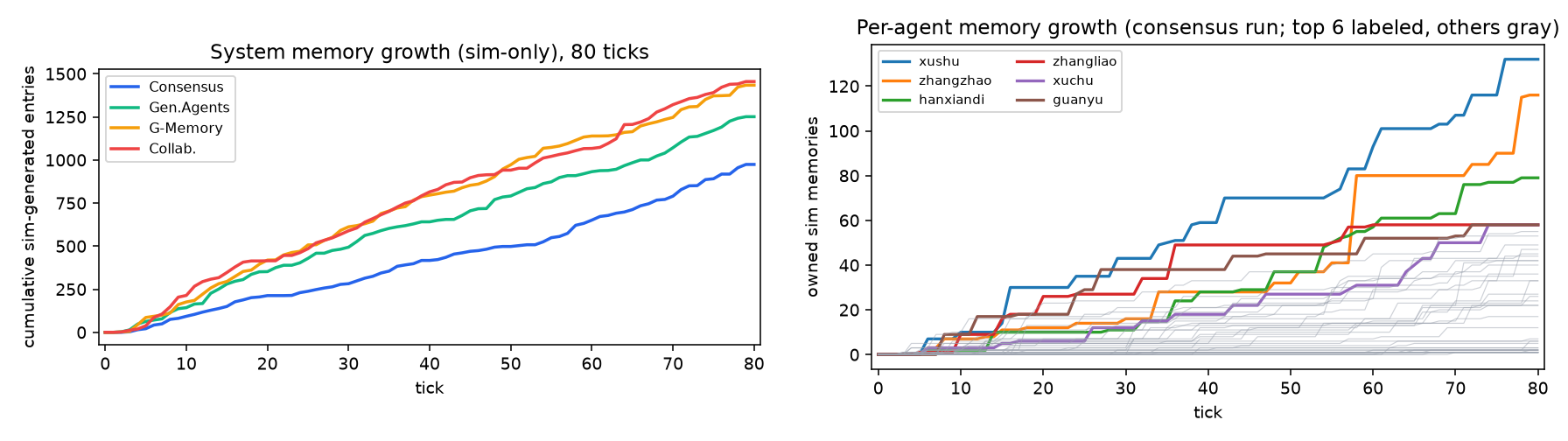}
\caption{Memory growth, Three Kingdoms, 80 rounds. Left: cumulative
simulation-written entries per round for all four backends, under the same
accounting as Table~\ref{tab:footprint} --- the consensus store is below
every baseline from the opening rounds and the gap widens, because each merge
removes an entry that would otherwise have been written and re-recorded by
later witnesses. Right: per-agent owned memories in the consensus run,
busiest six agents labelled. The other three worlds repeat the pattern
(Appendix~\ref{app:growth}).}
\label{fig:growth}
\end{figure}

\begin{figure}[t]
\centering
\includegraphics[width=0.86\textwidth]{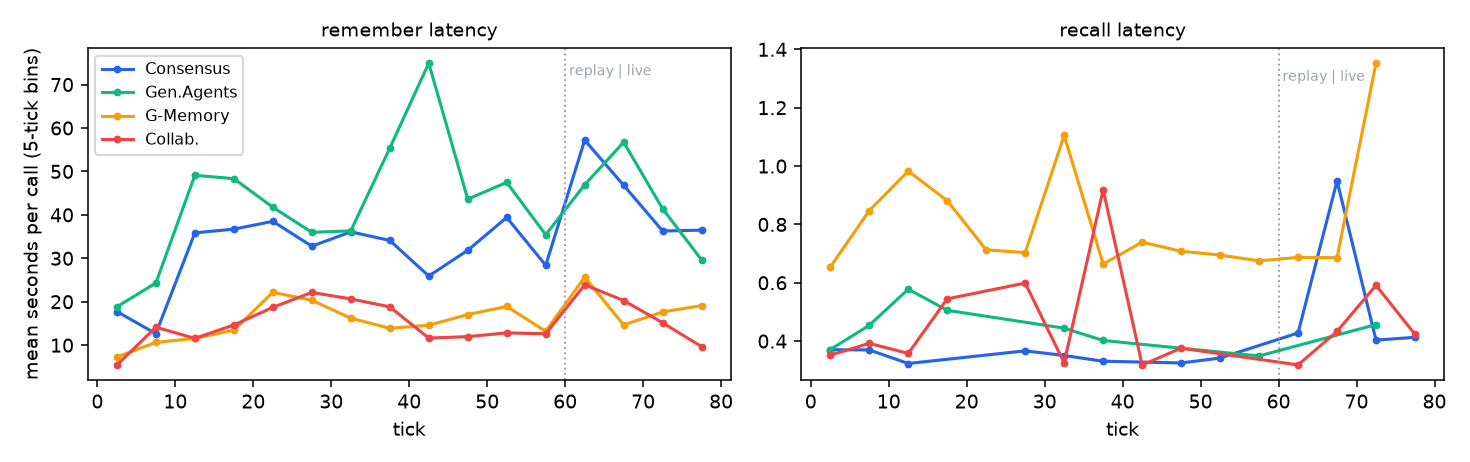}
\caption{Memory-operation latency, Three Kingdoms: mean seconds per
\code{remember} (left) and \code{recall} (right), 5-round bins. Writes cost
tens of seconds wherever the mechanism calls a model on the write path ---
consensus for the equivalence judge, Generative-Agents for importance and
reflection --- while reads stay under about a second for every backend,
because retrieval calls no model. The profile repeats in the other three
worlds (Appendix~\ref{app:growth}).}
\label{fig:latency}
\end{figure}

The separation of Table~\ref{tab:footprint} is not an end-state artefact: the
consensus curve is below every baseline from the opening rounds and the gap
widens rather than closing (Figure~\ref{fig:growth}), and the per-agent panel
shows the same mechanism from the agent's side --- a handful of principals
accumulate most of what a world learns. The latency profile
(Figure~\ref{fig:latency}) locates the cost of a memory design on its write
path, with reads uniformly cheap; both hold unchanged in the other three
worlds (Appendix~\ref{app:growth}).

\subsection{Structure compounds with horizon}
Sharing is not a fixed property but a function of how long the world has run:
Red Chamber, run at four horizons, goes \RcShChain{} at
rounds 10/40/60/80, and Russia--Ukraine \RuShChain{} at 10/20/40. Later
deposits meet a store that already holds their equivalents, so the compression
compounds rather than saturating early.

\subsection{Continuation quality}
\label{sec:quality}

\begin{figure}[t]
\centering
\includegraphics[width=0.94\textwidth]{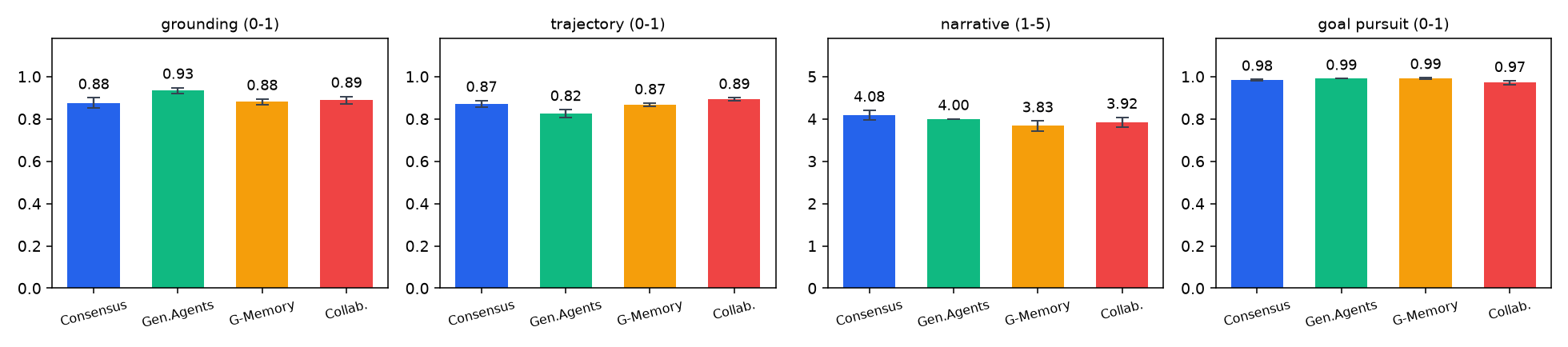}
\caption{Continuation quality --- Three Kingdoms, 80 rounds: grounding,
trajectory, narrative and goal pursuit, bars are means over three scorings
with $\pm1$ std whiskers. The first three panels are scored from the rendered
screenplay, the fourth from the event log. The other three worlds, and the
full table of means, are in Appendix~\ref{app:quality}.}
\label{fig:quality}
\end{figure}

Of the \QualityCells{} world-metric cells, consensus is best in \QualityLeads{}
and trails elsewhere by margins the size of the spread across three scorings of
the same text. We report this as a wash rather than a win: compression neither
buys judged quality nor costs it, and the case for consensus rests on
\S\ref{sec:structure}. Goal pursuit saturates at \GoalLo{}--\GoalHi{} for every
backend---the goal stack is in every prompt and only \GoallessLo{}--\GoallessHi{}
of actions are taken with no goal held---so it certifies the agent loop and
separates nothing. The wash is not a ceiling artefact
throughout: goal pursuit is ceiling-bound (\HeadGoal{} of range unused, spread
at most \SpreadGoal{} within a world) but narrative is not (\HeadNarr{} unused,
spread up to \SpreadNarr{}), so that rubric has room to separate the backends
and does not. The 40--80 round horizon remains a live alternative.

\subsection{The memory graph tracks the social graph}
\label{sec:social}
That conversing agents share more memories is what the merge rule is built to
produce, and alone would be a check that it runs. The finding is that the
\emph{partition} is recoverable from the store alone---the heatmap's block
structure reproduces the conversational communities without reading the
interaction log.
\begin{figure}[t]
\centering
\includegraphics[width=0.95\textwidth]{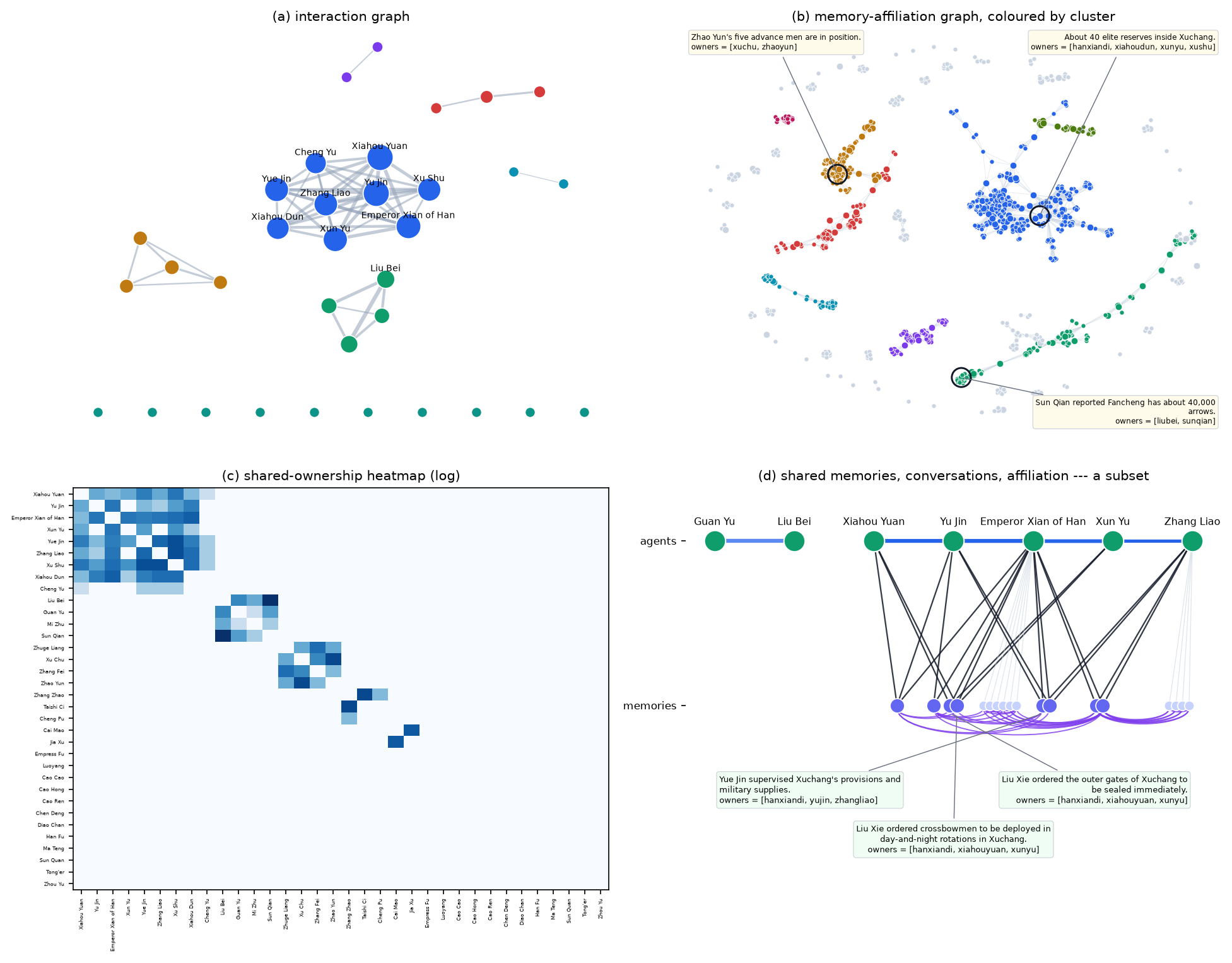}
\caption{The three layers of one run --- Three Kingdoms, 80 rounds.
(a)~Interaction graph: each node an agent, sized by degree, coloured by
conversational community; the court around the emperor, the Liu Bei camp and
the Jiangdong envoys separate without being told apart. (b)~Memory-affiliation
graph: each node one consensus memory, edges are affiliation, colour is
cluster; three memories from different clusters are circled with their text
and owner set --- a cluster is a plotline, and its memories' owner sets name
the agents living it. (c)~Shared-ownership heatmap: cell $(i,j)$ counts
memories owned by both agents (log scale); its block structure is (a)'s
communities seen from the store. (d)~A subset drawn in full: five conversing
agents (blue edges, weighted), the memories they own (dark lines and larger
nodes where a memory has several owners), affiliation arcs among those
memories (purple), and three memories labelled with text and owners ---
one record, several owners, is what consensus compression produces and what
every baseline lacks. The other worlds are in Appendix~\ref{app:case}; in
the HTML version these figures are interactive.}
\label{fig:case}
\end{figure}

\begin{figure}[t]
\centering
\includegraphics[width=0.84\textwidth]{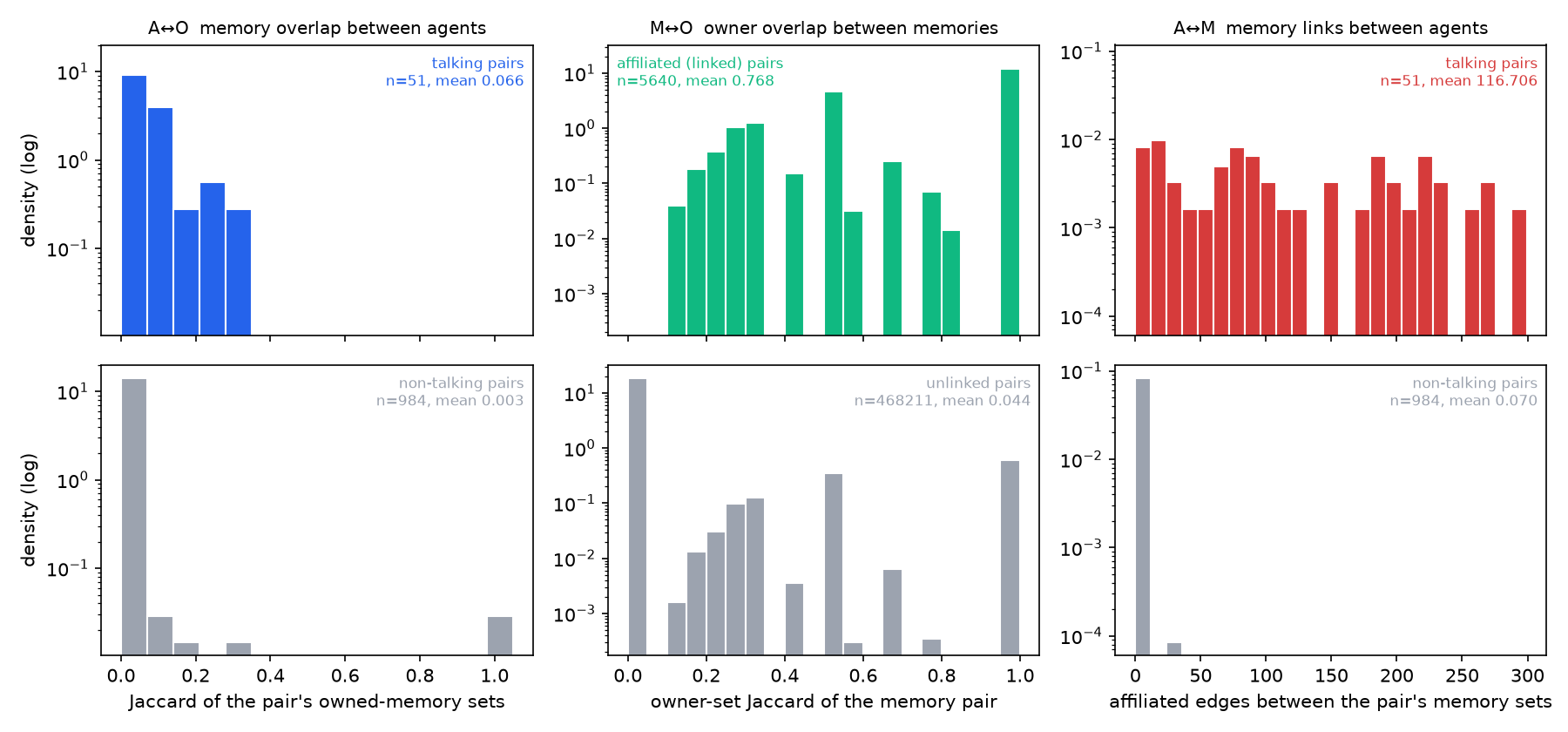}
\caption{The three pairwise relations, Three Kingdoms. Each panel contrasts
pairs that have the property with pairs that do not: agents who converse
against agents who do not, linked memories against unlinked ones, and
cross-agent memory links against the rest.}
\label{fig:relations}
\end{figure}

Three layers of the same run can be compared: who talked to whom, which memories
are affiliated, and who owns what (Figure~\ref{fig:case}). Agents who converse overlap in memory an
order of magnitude more than those who do not, linked memories share witnesses
far more often than unlinked ones, and cross-agent memory links run almost
exclusively between conversing pairs. The structure the mechanism produces is
therefore not incidental: the story's social organization can be read off the
memory substrate alone.

\subsection{Agents do not manage memory}
\label{sec:nomgmt}
Across all four backends and both models tested, agents issued zero calls to
every discretionary memory-management action---linking, forgetting, revising,
graph traversal by id---while calling \code{remember} and \code{recall}
constantly. This is why every structural mechanism in \S\ref{sec:consensus} runs
inside those two operations. Zero calls under one regime is not proof that
agent-side curation cannot work: we did not vary the actions' prominence in the
prompt, attach incentives, or run past 80 rounds, and each is an untested
alternative. What it establishes is the weaker claim---structure cannot be
\emph{assumed} to arrive from agent-side curation.

\subsection{Ablation}
\label{sec:ablation}

\begin{table}[t]
\centering
\caption{One factor at a time from the published configuration, consensus on
Three Kingdoms, 40 rounds. All four quality metrics here read the event log, so
they compare across cells rather than against Table~\ref{tab:quality}.}
\label{tab:ablation}
\resizebox{\textwidth}{!}{% generated
\begin{tabular}{llrrrrrrrr}
\toprule
Merge & Cache & Store & Sim & Shared & Linked & Grnd & Traj & Narr & Goal \\
\midrule
on & fifo \emph{(published)} & 6,590 & 539 & 19\% & 93\% & 0.92 & 0.89 & 4.17 & 0.97 \\
on & relevance & 6,674 & 623 & 21\% & 96\% & 0.93 & 0.83 & 4.33 & 0.99 \\
on & hybrid & 6,670 & 619 & 18\% & 96\% & 0.94 & 0.90 & 4.25 & 0.98 \\
\textbf{off} & fifo & 20,669 & 813 & 0\% & 94\% & 0.92 & 0.82 & 4.00 & 0.96 \\
\bottomrule
\end{tabular}
}
\end{table}

Disabling the merge multiplies the store by \AblFactor{}$\times$
(\AblOffEntries{} entries against \AblOnEntries{}) and takes sharing to exactly
zero: P1 measured against its own control rather than against a different
system. Two findings correct the paper's own framing. First, sharing and linking
are \emph{separable}: with the merge off, linking is still \AblOffAff{}, against
\AblOnAff{} published, because links come from the atomizer rather than from the
merge. Second, the short-term cache policy is not a research variable---
relevance and hybrid eviction leave every structural quantity where FIFO leaves
it, at 15--19\% more wall-clock. The quality columns do not separate on the merge
either. What the merge does cost is write latency
(Figure~\ref{fig:ablation_latency}): a deposit runs an extra judge call, so
\code{remember} takes about twice as long, while \code{recall} is unchanged.

\section{Discussion and limitations}
\label{sec:discussion}
Consensus compression buys structure and footprint at no measurable cost in
judged behaviour. What is new is not the components---atomization, an embedding
pre-filter and an LLM equivalence judge are each standard---but making ownership
a property of the record, not the store. Three costs follow, measured
in Appendix~\ref{app:cost}. Keeping the shorter of two equivalent texts is not the
neutral choice it looks: on the \MergeCov{} of merges whose discarded side the
log preserves, the surviving text drops a named character in \MergeLossPct{} of
cases and the ordering party specifically in \MergePrincPct{}, so the store
keeps the event and loses the chain of command.
Auto-expansion is the expensive one: recall appends every affiliated sibling the
agent owns, unranked and uncapped, returning \ExpAmpRange{} the rows---markedly
less on-query than the semantic hits, and a payload that grows with the very
merging that shrinks the store. Equivalence is decided by a single
unverified LLM call over five candidates, so a wrong merge is permanent; of the
\MergeRecov{} merges we can recover and check by hand, five conflate distinct
events.
Scale bounds all of it: the largest store is \StoreRowsMax{} rows, where a
vector index is not under strain, so the case here is footprint and
structure rather than retrieval feasibility, and merging roughly doubles write
latency (\S\ref{sec:ablation}). Every number comes from a single stochastic run.

\section{Conclusion}
Agentsensus treats a story world's memory as a single consensus store: equivalent
memories merge across witnesses, split memories link into a graph, and recall
walks that graph automatically. Across four worlds this yields the smallest
footprint at equal granularity, the only shared and linked memory structure
among four backends, and an emergent graph mirroring the story's social
structure, at judged quality within noise of the baselines. The broader lesson is narrower than it looks: agents
reliably used only \code{remember} and \code{recall}, so a design that waits for
agent-side curation gets no structure---at this horizon, under the one
prompting regime we tested.

\section*{Reproducibility Statement}
Everything needed to run the method is released, anonymised for review at
\url{https://anonymous.4open.science/r/agentsensus-562C/}: the framework,
all four
memory backends, the four scenarios with their source texts and information
carriers, and the scripts for every stage of the pipeline --- sedimentation,
the runs, the structure statistics, the four quality metrics, the ablation
grid, and the figures. \texttt{REPRODUCE.md} walks a clone through that
pipeline in order, with the command for each stage and its measured cost;
Hamlet at 40 rounds is the cheapest entry point that still shows every
structural effect we report. What is deliberately \emph{not} released is the
data a run produces --- stores, checkpoints, event logs --- which is large and
regenerable. Reproduction is of the method rather than of the numbers: every
stage calls an LLM, so a fresh run differs in its particulars. What should
reproduce is the shape of the result: the fewest entries at equal
granularity, sharing and linking exclusive to consensus, and sharing rising
with the horizon. The one cell we reran under identical settings landed
within 1\% of its original (\S\ref{sec:ablation}).

\section*{AI Use Statement}
Large language models appear in this work in two distinct roles, which we
separate here.

\paragraph{As the object of study.} The simulation is LLM-driven and the
continuation-quality metrics are LLM-judged. Both are part of the method
rather than of its preparation; the models, versions and prompts are
specified in \S\ref{sec:framework}, \S\ref{sec:setup} and
Appendix~\ref{app:impl}.

\paragraph{As a tool in preparing this paper.} An LLM-based coding assistant
was used to draft sections of the text, to write the analysis and
figure-generation scripts released with the code, and to run the post-hoc
measurements reported in Appendix~\ref{app:cost}. Every quantity we report is
produced by those scripts reading the event logs and stores on disk, and the
authors read the scripts and checked their output against the underlying
records; no number or claim in this paper rests on an assistant's assertion
alone. Where an assistant's initial analysis was wrong it was corrected
against the data, and the corrected measurement is what appears here. The
authors take full responsibility for the content of the paper, including any
remaining errors.

\IfFileExists{iclr2027_conference.bst}{\bibliographystyle{iclr2027_conference}}{}
\bibliography{references}

\appendix

\section{Implementation details}
\label{app:impl}

Everything in this section is shared by all four backends; only the long-term
store differs. It is collected here so \S\ref{sec:framework} can stay on the
memory mechanism.

\paragraph{Agent kinds and scheduling.} Characters are the only schedulable
agents. Environments and information carriers are passive: the kernel never
gives them a turn, so an action aimed at one (\code{act\_on}, \code{read})
is resolved synchronously inside the acting character's own apply step rather
than on the target's next tick. A character that is dead or absent by the end
of the source material still gets an agent, so its memories stay in the store
and remain recallable by the living, but it is never scheduled.

\paragraph{The round barrier.} A round proceeds in two phases. Every awake
agent first builds its view and decides against the world state as it stood at
the barrier; only then are the chosen actions applied. Effects therefore become
visible in the following round, and the outcome of a round does not depend on
the order agents were stepped in, which is what makes a run reproducible from
its event log.

\paragraph{Messages and distance.} \code{say} and \code{gesture} are the two
asynchronous actions: they may only target a character, and they deliver into
the recipient's inbox, visible from the next round. The kernel holds the
exchange as a conversation thread that either party can re-read with
\code{read\_thread}. Delivery is delayed in proportion to the distance between
speaker and recipient, so a letter across the map takes rounds to arrive.

\paragraph{The action repertoire.} Agents choose one action per round from
twenty-four: twenty-two synchronous and the two asynchronous ones above. Six
of the twenty-four are discretionary memory management ---
\code{forget}, \code{revise\_memory}, \code{add/remove/set\_affiliated} and
\code{get\_affiliated} --- and were never called (\S\ref{sec:nomgmt},
Appendix~\ref{app:actions}). Figure~\ref{fig:framework} names the six that act
on the world or on another agent; the rest are cognitive (\code{think},
\code{conclude}, \code{wait}, \code{noop}), memory (\code{remember},
\code{recall}) or short-term bookkeeping.

\paragraph{Short-term memory.} Each character carries a short-term memory of
three parts, and exactly these three are serialised into the prompt each round:
a \emph{goal stack} (index zero the most fundamental, maintained by
\code{push\_goal}, \code{pop\_goal} and \code{replace\_goal}); a
\emph{status register} of key--value pairs with a private subset, maintained by
\code{update\_status} and \code{remove\_status}; and a \emph{cache} of the
twenty most recent \code{(action, result)} pairs. When the cache is full the
evicted pair is chosen by one of three policies: \code{fifo} drops the oldest
(the published configuration), \code{relevance} drops the existing pair least
similar to the incoming one by embedding cosine, and \code{hybrid} scores
$\alpha\cdot\text{recency} + (1-\alpha)\cdot\text{relevance}$ with
$\alpha = 0.5$. The pair being appended is always kept. \S\ref{sec:ablation}
varies this policy and finds it changes no structural quantity.

\section{What the mechanism costs}
\label{app:cost}

Table~\ref{tab:cost} collects the measurements behind \S\ref{sec:discussion}.
All of it is arithmetic over event logs and final stores already on disk
(\code{experiments/review\_stats.py} and \code{experiments/merge\_loss.py}); no rerun is involved.

\paragraph{Auto-expansion.} A \code{recall} runs a semantic query for
\ExpBaseK{} rows the agent owns, then follows each hit's affiliation edges one
hop and appends every linked row the agent also owns. The appended rows never
compete on similarity and there is no cap, so what returns is
\ExpAmpRange{} the semantic result. The same path serves \code{read}, since
\code{recall} is a one-line call into it, so information carriers are read
through the same expansion. Affiliation links exist because atomization splits a
compound deposit into self-contained statements; expansion reassembles them at
read time. That also means a row accumulates siblings as it absorbs deposits:
in Red Chamber the median row carries \AffRcLow{} affiliated siblings at one
owner and \AffRcHigh{} at four or more, so the read-side payload grows with the
same merging that shrinks the store. Russia--Ukraine is the control: its
institutional deposits rarely split, so it has almost no siblings to follow and
almost no expansion. The principled repair is to rank affiliated rows into the
candidate pool and truncate, rather than append them; we have not run that, nor
run the mechanism with expansion off, so the \emph{benefit} of expansion is
unmeasured here even though its cost is not.

\paragraph{Keeping the shorter text.} When two items are judged equivalent the
store keeps whichever text is shorter. That rule is not neutral. The judge's
criterion is that the two describe the same event, and among descriptions of one
event the shorter is systematically the one that omits the principal: ``X ordered
Y to do Z'' is necessarily longer than ``Y did Z''. Measuring this needs both
input texts, which the log preserves only when the surviving text changed or the
deposit was a single atom---\MergeRecov{} of \MergeTotal{} merges (\MergeCov{}).
On that subset the loss is small in characters (median \MergeChars{}, at most
\MergeCharsMax{}) but not in content: in \MergeLossN{} of \MergeRecov{} merges
(\MergeLossPct{}; \MergeLossRange{} across worlds) the surviving text drops a
named character the discarded text carried, and in \MergePrincN{}
(\MergePrincPct{}) the dropped name is the party that ordered, authorized or
dispatched the act. Names resolve through the scenario registry, so two aliases
of one person do not count as a loss. Two cases stand for the rest: ``Emperor
Xian ordered Xiahou Yuan to ride out from Xuchang with three hundred light
cavalry'' survives as ``Xiahou Yuan rode out from Xuchang with three hundred
light cavalry'', and ``Zelenskyy authorized the SBU and GUR to nominate liaison
officers'' as ``SBU and GUR will nominate liaison officers''. The event is kept;
the chain of command is not. The subset is not a random sample---it is the
merges the log happens to expose---so the rate is an estimate over observable
merges rather than a population figure. Set against this, the store shows no
erosion in length with depth: Red Chamber's median row grows from \LenRcLow{}
to \LenRcHigh{} characters between one and four-or-more owners, which is why
counting characters misses the defect entirely. The rule is one line in the
merge path and the repair is equally small---keep the longer text, or the
union---costing \MergeCharsMax{} characters in the worst case seen here. We did
not make that change: every run reported here was produced under the shipped
rule, and we did not rerun. The defect is characterized, not fixed.

\paragraph{Two caveats on the relevance columns.} Whether a returned row
contains the query string is a keyword proxy: it counts a row as off-query when
it is merely worded differently, so both columns understate relevance and only
their \emph{ratio} carries weight. It fails outright on Hamlet, whose queries
are English clauses rather than names---hence \(0\%\) in both columns, which
should be read as ``proxy inapplicable'', not ``nothing relevant''. Hamlet also
issued only three recalls in forty rounds, too few to carry an estimate.

\begin{table}[h]
\centering
\caption{What the mechanism costs, per world. \emph{recall}: number of calls,
median semantic hits plus rows added by one-hop expansion, the resulting
amplification, and how often the query string appears in each group.
\emph{affiliated}: median siblings carried by a row at one owner versus four or
more. \emph{keep-shorter}: merges whose discarded side the log
preserves, over all merges; the median characters discarded; the share of those
merges whose surviving text loses a named character; and the share where the
lost name is the party that ordered the act.}
\label{tab:cost}
\resizebox{\textwidth}{!}{% generated
\begin{tabular}{lrrlrrrrrrrr}
\toprule
& & \multicolumn{5}{c}{\emph{recall}} & \emph{affiliated} & \multicolumn{4}{c}{\emph{keep-shorter}} \\
\cmidrule(lr){3-7}\cmidrule(lr){8-8}\cmidrule(lr){9-12}
World & rows & $n$ & hits+exp. & amp. & q\,in hits & q\,in exp. & 1$\to$4+ own. & seen & chars & $-$name & $-$principal \\
\midrule
Three Kingdoms & 7,025 & 51 & 5+27 & 6.6$\times$ & 25\% & 9\% & 8$\to$13 & 60/447 & 4 & 30\% & 15\% \\
Red Chamber & 6,944 & 19 & 5+35 & 8.7$\times$ & 32\% & 9\% & 4$\to$19 & 43/267 & 5 & 37\% & 7\% \\
Russia--Ukraine & 2,347 & 41 & 5+1 & 1.6$\times$ & 10\% & 0\% & 7$\to$1 & 37/314 & 20 & 22\% & 14\% \\
Hamlet & 1,265 & 3 & 5+40 & 8.8$\times$ & 0\% & 0\% & 2$\to$26 & 7/74 & 9 & 29\% & 14\% \\
\bottomrule
\end{tabular}
}
\end{table}

\section{What agents actually call}
\label{app:actions}

Table~\ref{tab:actions} is the census behind \S\ref{sec:nomgmt}: every action
every backend invoked, counted from the event logs of each world's final
stage. Two facts are visible at once. \code{remember} and \code{recall} are
called constantly, and the six discretionary memory-management actions ---
\code{forget}, \code{revise\_memory}, \code{add/remove/set\_affiliated},
\code{get\_affiliated} --- do not appear at all, in any world, for any
backend, which is why they are absent from the table rather than shown as rows
of zeros. The actions were documented in the same skill text as the two that
are used, with worked examples and an id-free query interface, and were
available in every round of every run.

\begin{table}[h]
\centering
\caption{Actions called per backend, by world, over each world's final stage.
Counted from the event logs. Discretionary memory-management actions are
omitted because every count is zero.}
\label{tab:actions}
\resizebox{0.75\textwidth}{!}{% generated
\begin{tabular}{lrrrr}
\toprule
Action & Consensus & Gen.\ Agents & G-Memory & Collaborative \\
\midrule
\multicolumn{5}{l}{\emph{Three Kingdoms}} \\
\quad\texttt{say} & 66 & 50 & 48 & 61 \\
\quad\texttt{read\_thread} & 123 & 110 & 83 & 121 \\
\quad\texttt{observe} & 28 & 34 & 31 & 37 \\
\quad\texttt{move} & 2 & 1 & 4 & 1 \\
\quad\texttt{act\_on} & 11 & 14 & 23 & 22 \\
\quad\texttt{read} & 8 & 12 & 36 & 9 \\
\quad\texttt{think} & 3 & 1 & 3 & 2 \\
\quad\texttt{conclude} & 0 & 0 & 2 & 0 \\
\quad\texttt{push\_goal} & 104 & 117 & 77 & 110 \\
\quad\texttt{pop\_goal} & 25 & 20 & 24 & 30 \\
\quad\texttt{replace\_goal} & 0 & 0 & 1 & 1 \\
\quad\texttt{update\_status} & 3 & 0 & 1 & 0 \\
\quad\texttt{remember} & 48 & 48 & 41 & 52 \\
\quad\texttt{recall} & 12 & 2 & 9 & 10 \\
\quad\texttt{wait} & 52 & 64 & 69 & 51 \\
\midrule
\multicolumn{5}{l}{\emph{Red Chamber}} \\
\quad\texttt{say} & 89 & 80 & 84 & 72 \\
\quad\texttt{read\_thread} & 125 & 119 & 131 & 93 \\
\quad\texttt{observe} & 28 & 26 & 24 & 25 \\
\quad\texttt{move} & 8 & 7 & 11 & 12 \\
\quad\texttt{act\_on} & 2 & 2 & 5 & 6 \\
\quad\texttt{read} & 2 & 0 & 0 & 1 \\
\quad\texttt{push\_goal} & 113 & 109 & 115 & 85 \\
\quad\texttt{pop\_goal} & 29 & 29 & 30 & 20 \\
\quad\texttt{replace\_goal} & 0 & 0 & 0 & 1 \\
\quad\texttt{update\_status} & 1 & 0 & 0 & 0 \\
\quad\texttt{remember} & 46 & 38 & 43 & 36 \\
\quad\texttt{recall} & 3 & 0 & 4 & 1 \\
\quad\texttt{wait} & 36 & 32 & 25 & 35 \\
\midrule
\multicolumn{5}{l}{\emph{Russia--Ukraine}} \\
\quad\texttt{say} & 144 & 136 & 151 & 156 \\
\quad\texttt{read\_thread} & 253 & 234 & 241 & 263 \\
\quad\texttt{observe} & 20 & 9 & 16 & 11 \\
\quad\texttt{act\_on} & 6 & 5 & 9 & 16 \\
\quad\texttt{read} & 7 & 11 & 18 & 10 \\
\quad\texttt{think} & 0 & 3 & 2 & 1 \\
\quad\texttt{conclude} & 2 & 0 & 7 & 0 \\
\quad\texttt{push\_goal} & 177 & 185 & 195 & 143 \\
\quad\texttt{pop\_goal} & 36 & 77 & 45 & 40 \\
\quad\texttt{replace\_goal} & 1 & 1 & 3 & 0 \\
\quad\texttt{update\_status} & 2 & 1 & 0 & 1 \\
\quad\texttt{remember} & 62 & 84 & 76 & 78 \\
\quad\texttt{recall} & 4 & 6 & 11 & 8 \\
\quad\texttt{wait} & 50 & 39 & 34 & 46 \\
\midrule
\multicolumn{5}{l}{\emph{Hamlet}} \\
\quad\texttt{say} & 30 & 29 & 28 & 24 \\
\quad\texttt{read\_thread} & 32 & 36 & 42 & 31 \\
\quad\texttt{observe} & 5 & 13 & 17 & 8 \\
\quad\texttt{move} & 1 & 0 & 2 & 1 \\
\quad\texttt{act\_on} & 1 & 0 & 0 & 1 \\
\quad\texttt{push\_goal} & 24 & 38 & 36 & 28 \\
\quad\texttt{pop\_goal} & 3 & 10 & 4 & 13 \\
\quad\texttt{update\_status} & 0 & 1 & 0 & 0 \\
\quad\texttt{remember} & 8 & 12 & 11 & 14 \\
\quad\texttt{wait} & 6 & 4 & 7 & 4 \\
\bottomrule
\end{tabular}
}
\end{table}

\section{Transcript excerpts and screenplay samples}
\label{app:excerpts}

\subsection{One message per backend}
One deduplicated \code{say} message per backend, from each world's final
stage, chosen by length so a block fits here; only the recipient list is
abridged. The two Chinese worlds are given in English with the original
beneath; the two English worlds are given as written. The point of the
comparison is not prose quality --- \S\ref{sec:quality} measures that and
finds the backends within error bars of one another --- but what a memory
design looks like from the outside.

% Appendix A.2 -- one deduplicated say message per backend, from each world's
% final stage. Ported from the HTML paper (build_paper_academic.py), which is
% where the excerpts were selected; Chinese originals render via CJKutf8.

\newcommand{\exq}[1]{\begin{quote}\small #1\end{quote}}

\subsection*{Three Kingdoms (80 rounds, Chinese)}

\exq{\textbf{consensus}\quad[r61] Cheng Yu \zh{程昱} $\to$ Yue Jin \zh{乐进}\\
Yue Jin: regarding tonight's midnight collection of Xu Shu's mother's letter
east of Xuchang --- which timing now stands? Still midnight, or moved to the
third quarter after noon tomorrow at the post-house? Report back at once.\\
\textcolor{gray}{\zh{乐进：今夜子时于许昌城东取徐母家书之事，今以何时为准？仍按子时取回，还是改为明日午后三刻于馆驿交接？请速回禀。}}}

\exq{\textbf{generative-agents}\quad[r61] Mi Zhu \zh{糜竺} $\to$ Liu Bei \zh{刘备}, Guan Yu \zh{关羽}, Sun Qian \zh{孙乾}\\
Reporting to my lord: the ruling is received. The force stands at three
hundred; three days' rations drawn from Fancheng are approved; I will issue
the supplies at once and have the Fancheng grain officer seal and register
them\ldots\\
\textcolor{gray}{\zh{回禀主公：承受圣裁。兵数仍定为三百，准用樊城抽拨之三日口粮；我即刻领发粮草并督樊城粮官封存登记……}}}

\exq{\textbf{g-memory}\quad[r62] Xiahou Yuan \zh{夏侯渊} $\to$ Xun Yu \zh{荀彧}\\
Lord Xun: Yu Jin and Yue Jin have carried the registers to the yamen; I have
also compiled the scouts' summaries and the camp inspection books\ldots{} can
it be consolidated into a memorial to His Majesty within three quarters, as
agreed?\\
\textcolor{gray}{\zh{荀公：于禁、乐进已携册赴府衙呈上，臣亦已将斥候回报摘要与各营点验簿整理成册……能否按议于三刻内汇整为摘要上奏陛下？}}}

\exq{\textbf{collaborative}\quad[r61] Xu Chu \zh{许褚} $\to$ Guan Yu \zh{关羽}, Zhuge Liang \zh{诸葛亮}, Zhao Yun \zh{赵云}\\
Reconnaissance of Mount Bailang is complete. Observing along the ridge, no
main enemy force was seen --- only scattered scout riders and a few suspicious
tents and campfires. Signals are set; returning by the same route, arrival
expected within the hour.\\
\textcolor{gray}{\zh{白狼山侦察已毕。沿山脊侦望，未见敌军主力，仅见零星侦骑与少数可疑营帐、营火。已设回报信号，现按原路返营，预计一小时内抵达。}}}

\subsection*{Red Chamber (80 rounds, Chinese)}

\exq{\textbf{consensus}\quad[r60] Jia Baoyu \zh{贾宝玉} $\to$ Grandmother Jia \zh{贾母}, Jia Zhen \zh{贾珍}, Wang Xifeng \zh{王熙凤}\\
Grandmother, Jia Zhen, Sister Feng, Baochai --- I am waiting here. Does
Grandmother wish me to come up and sit beside her now, or to wait until Miss
Lin arrives and be seated together?\\
\textcolor{gray}{\zh{外祖母、贾珍、凤姐、宝钗，晚辈已在此恭候。请问外祖母现在是要我先近前到贾母身侧就座陪侍，还是等林姑娘到齐再并列就座？}}}

\exq{\textbf{generative-agents}\quad[r60] Jia Lian \zh{贾琏} $\to$ Ban'er \zh{板儿}, You-shi \zh{尤氏}\\
Ban'er, state your name and Granny's, your business, and whether you carry a
letter of introduction or any proof of identity; You-shi, have a page bring
Ban'er and Granny Liu to the hall to wait, and let them not wander\ldots\\
\textcolor{gray}{\zh{板儿，请将你与姥姥的姓名、来意及是否带有介绍信或能证明身份的凭证一并禀明；尤氏，请差小厮将板儿与刘姥姥领到厅上等候，不可四处走动……}}}

\exq{\textbf{g-memory}\quad[r60] Jia Tanchun \zh{贾探春} $\to$ Li Wan \zh{李纨}\\
Sister Li, as you asked, I have offered the club a model verse at the table
(``the crab-apple newly opened, dew holding its scent\ldots''), Xiangyun has
answered it, and the others are responding in turn\ldots{} as for this month's
host, Xichun is willing to take a turn.\\
\textcolor{gray}{\zh{李姐姐，承你所托，探春已于席间为社中示范一首（海棠初绽露凝香……）且湘云已和唱，诸位正依次应和……关于本月轮主，惜春愿任一任。}}}

\exq{\textbf{collaborative}\quad[r60] Qingwen \zh{晴雯} $\to$ Jia Yun \zh{贾芸}, Xiaohong \zh{小红}\\
Jia Yun, were you just now holding a white embroidered handkerchief with a
word stitched in the corner? Xiaohong says it is hers. If you have it, produce
it for her to see; if not, say so plainly.\\
\textcolor{gray}{\zh{贾芸，你刚才可曾拿着一方白底绣花、角上有绣字的绢子？小红说那是她的，若你有请拿出来让小红看；若没有请直言。}}}

\subsection*{Russia--Ukraine (40 rounds, English)}

\exq{\textbf{consensus}\quad[r24] un $\to$ guterres\\
``Rosemary DiCarlo, please begin your 5-minute brief now and name the DPA
drafting lead for the Security Council text.''}

\exq{\textbf{generative-agents}\quad[r24] iaea $\to$ grossi\\
``Director General IAEA requests immediate confirmation of the current status
at Zaporizhzhia NPP. Please secure safe, unimpeded access for an IAEA expert
mission, coordinate with Energoatom\ldots''}

\exq{\textbf{g-memory}\quad[r24] zaluzhnyi $\to$ radakin\\
``Brigadier Foster --- I am transmitting Colonel Andriy Kovalenko's secure
contact details and the prioritized spreadsheet to you and the UK team now via
the agreed secure channel\ldots''}

\exq{\textbf{collaborative}\quad[r33] un $\to$ guterres\\
``Secretary-General, DPO has not yet delivered its update. Shall OCHA proceed
now with its five-minute update on civilian needs and corridor routes, or wait
a moment longer for DPO?''}

{\small These are English-language excerpts, but not every message in this
world was English: 12\% of the run's memories are Chinese and 2\% Russian
(Appendix~\ref{app:language}), the latter because agents adopted the language
of the character they were playing.}

\subsection*{Hamlet (40 rounds, English)}

\exq{\textbf{consensus}\quad[r24] fortinbras $\to$ marcellus\\
``Marcellus, report the current troop and weapon counts on the battlements,
and prepare to escort me to the Great Hall so we can inventory arms and gather
grain.''}

\exq{\textbf{generative-agents}\quad[r20] marcellus $\to$ barnardo\\
``Barnardo --- go wake Horatio at once and tell him to come to the battlements
immediately. Do not go alone; wait here until he arrives.''}

\exq{\textbf{g-memory}\quad[r28] barnardo $\to$ horatio\\
``Horatio, Marcellus and I just saw a ghost on the battlements.''}

\exq{\textbf{collaborative}\quad[r20] first\_player $\to$ rosencrantz\\
``Thank you, Rosencrantz. I'm glad the play is engaging. Watch King Claudius
closely and tell me if his countenance changes during the performance.''}

\paragraph{Reading.} Three things hold across all four worlds. \emph{The
register is set by the action repertoire, not by the memory design}: every
backend drifts toward administrative correspondence --- requests,
confirmations, rosters, timings --- because what an agent can do is speak, set
goals and report; Red Chamber's poetry club and Hamlet's battlements are
pulled into the same idiom as Three Kingdoms's supply trains. \emph{The worlds
differ more than the backends do}: the institutional world produces long
procedural messages, the chamber drama short ones, the household world sits
between with etiquette carrying most of the content. What does track the
memory design is the \emph{direction of reference}. Consensus lines
characteristically ask which branch of an already-shared arrangement now
applies --- Cheng Yu asks whether the pickup still stands at midnight
\emph{or} has moved to the post-house; Baoyu asks whether to sit now \emph{or}
wait; the UN chair asks a named official to begin \emph{the} brief. The
speaker treats the prior arrangement as a record both sides hold and asks only
for the delta. Baseline lines more often re-establish the arrangement before
acting on it --- restating troop counts, rations and identities the other
party already holds --- which is what an agent does when it cannot assume the
other party's copy matches its own. This is an observation on selected
excerpts, not a measurement; \S\ref{sec:quality} puts the backends within each
other's error bars, and we claim no more than that.

\subsection{The rendered screenplays}
\label{app:screenplays}

The continuation-quality judge of \S\ref{sec:quality} reads a screenplay
rendered from each run's event log: beats are grouped into scenes by place and
stretch of time, each scene is dramatized in one grounded pass, and nothing
outside the log may appear --- the cast, the location and every action are
constrained to what the run produced, while the wording must be rewritten
rather than copied. Table~\ref{tab:screenplays} gives each screenplay's
geometry; the full texts (176 scenes) are in the supplementary material, and
two sample scenes per world follow, the Chinese worlds with the
source-language rendering beneath the English --- both produced from the same
beats in one pass each, not by translating one into the other.

\begin{table}[h]
\centering
\caption{The four screenplays. ``Beats'' are the dramatizable events after
deduplication; ``scenes'' the groupings by place and stretch of time that each
become one render call. Length is the rendered markdown; the Chinese worlds
carry two renderings of the same beats.}
\label{tab:screenplays}
\resizebox{0.9\textwidth}{!}{% generated
\begin{tabular}{lrrrrrrr}
\toprule
World & Rounds & Scenes & Beats & Speakers & Places & English & Source lang. \\
\midrule
Three Kingdoms & 0--79 & 54 & 312 & 33 & 24 & 206k & 62k (zh) \\
Red Chamber & 0--79 & 62 & 390 & 34 & 26 & 98k & 30k (zh) \\
Russia--Ukraine & 0--39 & 48 & 334 & 47 & 20 & 313k & -- (en) \\
Hamlet & 0--39 & 12 & 117 & 16 & 6 & 49k & -- (en) \\
\bottomrule
\end{tabular}
}
\end{table}

% generated -- sample scenes from the rendered screenplays
\subsection*{Three Kingdoms}
\paragraph{Scene 27 · Xiangyang · rounds 25–43}\mbox{}\\[2pt]
\begin{zhblock}\small\setlength{\parskip}{3pt}
Jia Xu:
Where is Zhang Xiu now?
I will see him and urge him to join Cao Cao.
\par Cai Mao:
Zhang Xiu's situation is unclear.
I do not know his exact whereabouts.
I can send men secretly to learn his movements.
Or order gatekeepers to inquire; I will report back.
\par Jia Xu:
Send men now to secretly learn Zhang Xiu's movements.
Report to me at once if anything stirs.
\par Cai Mao:
(At Xiangyang, Cai Mao dispatches two spies. They go to Zhang Xiu's likely locations to learn his movements. He orders gatekeepers and outer-city scouts to search in secret. Any stir must be reported to Jia Xu immediately.)
\par Cai Mao:
As you ordered, I have sent two spies.
I have ordered gatekeepers to search secretly.
Outer-city scouts have been ordered as well.
Any hint of movement will be reported to you at once.
\par Jia Xu:
If the spies learn Zhang Xiu's location, report at once.
Also secretly find a safe place to meet.
Prepare small gifts to show sincerity.
I will go in person to persuade him.
\par Cai Mao:
Your instruction is noted; I will comply.
Spies were told to report at once if they find Zhang Xiu.
I will have a secret meeting place arranged.
Small gifts will be prepared to show sincerity.
We will report to you immediately once there is a lead.
\par\textcolor{gray}{\rule{0.3\textwidth}{0.4pt}}\par
{\color{gray} 贾诩:
张绣如今在哪里？我想觐见他，劝他归附曹操。你知道他近在何处吗？
\par 蔡瑁:
张绣近况不详，我目前未得其确切所在。
若公欲觐见，我可即刻派人暗中探问，或令门吏查问其行踪，得报即来禀。
\par 贾诩:
立刻派人暗中探问张绣行踪，得报速来禀。
若有风吹草动，立即传报我。
\par 蔡瑁:
(派遣两名暗探前往张绣可能去处打探行踪。 他并令馆舍门吏与外城探子暗中查访，务须得报速来禀。 若有风吹草动，立刻将消息传报贾诩。)
\par 蔡瑁:
贾公所令已遵，已派两名暗探前往张绣可能去处打探。
馆舍门吏与外城探子亦被令暗中查访，若有风声必即刻禀报于公。
\par 贾诩:
若暗探得知张绣所在，务必速来禀报我。
并先暗中探明一处安全会面地点，备好薄礼以示诚意。
俟我亲往时，再行劝说。
\par 蔡瑁:
贾公所言，遵命。
已嘱暗探得张绣所在必速禀报，并即令暗中物色安全会面之地，备薄礼以示诚意。
若有着落，必即刻禀报于公。}
\end{zhblock}
\paragraph{Scene 47 · Xinye · rounds 65–70 (excerpt)}\mbox{}\\[2pt]
\begin{zhblock}\small\setlength{\parskip}{3pt}
Xu Chu:
Report received. I confirm: first, we will launch the night raid per the plan, but only after ten covering troops are in place and confirmed on site by cover leader Zhang Yong or by me.
Second: the start signal will be either Zhang Yong or I clearing our throats low three times; the interception team will move once the covers are in place and Zhang Yong or I give those three throat-clears; the retreat signal is three short whistle blasts and the word "Back"; rally at South River Bridge.
Third: current status — Zhao Yun's five vanguard are in position; he has twelve organized for interception and cover; his main nineteen wait under the big locust tree; Zhang Fei reports three hundred elite ready.
Fourth: orders — Zhang Fei, immediately form your three hundred per prior disposition and do not advance early; Zhao Yun, when cover is confirmed send your twelve as planned. Master Zhuge, please confirm you agree to this timing and these signals.
Fifth: estimate — the ten covering troops should arrive within about fifteen minutes; when they are in I will report the exact minute and call for final confirmation. Speak objections now, otherwise reply "Agree".
\par Zhang Fei:
I agree. Since Zhao Yun has twelve organized and Xu Chu's ten covers should be ready in about fifteen minutes, I take this moment as the start point and will launch in fifteen minutes.
The start signal will be Zhang Yong clearing his throat low three times; the retreat signal will be three short whistles and the word "Back".
I will lead three hundred elite in two columns, strike fast to seize the vanguard and avoid prolonged fighting; the entire night raid will not exceed thirty minutes; rally at South River Bridge.
Report any changes immediately, otherwise prepare to move at the appointed time.
\par\textit{[\,scene continues\,]}
\par\textcolor{gray}{\rule{0.3\textwidth}{0.4pt}}\par
{\color{gray} 许褚:
各位，我的回报已收到，现将结论与确认说明。
我方同意按既定预案发动夜袭，但须有先决条件。
先决条件为十名掩护精兵确到位，且由掩护首领张勇或我当场口头或目视确认。
起点信号为我或张勇低声清嗓三次。
截断队在掩护到位并经张勇或我低声三次确认后，由先遣引导出发。
撤退信号为同时吹三短哨并口令“回”，撤退集合地为南河桥。
现状说明：赵云先遣五人已到位，赵云已整编十二人截断与掩护，主力十九人在大槐树下待命。
张飞已报三百精锐待命。
命令：张飞按既定编制即刻整编三百精锐，切勿先发；赵云按既定在掩护确认后即令十二人出发。
十名掩护约在今刻起十五分钟内到位，届时我将回报确切整点分钟并发起最终确认，若无异议请回“同意”。
\par 张飞:
吾同意。
既然赵云已整编十二人，而许褚所派十名掩护约十五分钟后到位，吾以此刻为起算点。
十五分钟后发起夜袭，出发信号由掩护首领张勇低声清嗓三次。
撤退信号为三短哨并口令“回”。
吾率三百精锐分两路突击，速取先锋，不与敌缠斗。
夜袭全程不逾三十分钟，撤退集合地为南河桥。
若有变动速报，否则即按此时刻准备发动。
\par 诸葛亮:
张飞，我已收到你的回报，知你可率三百精锐分两路突击。
依许褚与赵云所言，今定约十五分钟后，在许褚与赵云双方确认到位并由许褚核实情报后即刻发动。
撤退集合地为南河桥，出发信号为掩护首领低声清嗓三次。
撤退信号为我方与赵云同时吹三短哨并口令“回”。
夜袭以速取速归为要，切勿缠斗。
请立刻整编出发人马，整编完毕请回报“同意”，以便我方计时。
\par 诸葛亮:
赵云、许褚、张飞，我承赵云所报，闻十二名截断掩护精兵已整编，主力在外待命。
今命赵云即刻入城，将侦得情报私呈许褚或掩护队长张勇，并回报可抽调之人数与进城时间。
许褚请如期核实，并于十名掩护到位时回报“到位”。
张飞继续整编，整编完毕请回报“同意”。
待赵云与许褚互证到位并由许褚核实情报后，自许褚回报之刻起约十五分钟即刻发动。
撤退集合地为南河桥，出发信号为掩护首领低声清嗓三次。
撤退信号为我方与赵云同时吹三短哨并口令“回”。
三方速回。
\par 诸葛亮:
赵云、许褚、张飞，今定如下执行程序。
十名掩护精兵一旦到位请立即回报“到位”，并由你或掩护队长张勇当场核验赵云所私呈之情报。
速将侦得情报私呈许褚或张勇，并明确此次可抽调执行截断与掩护的人数。
此前报十二名，请再确认可用人数。
回报“同意”以示三百精锐已整编完毕并按既定两路突击准备出发。
三方到位且许褚回报“到位”后，自该回报之刻起约十五分钟即刻发动。
撤退集合地为南河桥，出发信号与撤退信号照既定执行。
速回。
\par 赵云:
许褚、张飞、诸葛亮：赵云汇报并评估。
诸葛亮与张飞之分工可行，我方已整编十二人截断与掩护，先遣五人已到位，余十九人主力待命。
吾同意按既定预案发动夜袭，前提为许褚所派十名掩护确到并经许褚或掩护首领张勇当场确认。
请许褚在十人到位时立刻回报“到位”并报出确切整点分钟以便我方同步出发。
到位确认后，我方十二人将在先遣引导下，并经掩护或许褚低声清嗓三次为起点信号出发。
撤退集合地为南河桥，撤退信号为三短哨并口令“回”。
入城后我方先將侦得情报私呈张勇或许褚，地点为南门东侧旧仓库后廊。
张飞请继续三百精锐待命且勿先发；诸葛亮请再确认同意或速言异议。
赵云恭候回报。}
\end{zhblock}
\subsection*{Red Chamber}
\paragraph{Scene 5 · Jia Mansion · rounds 2–13}\mbox{}\\[2pt]
\begin{zhblock}\small\setlength{\parskip}{3pt}
Mingyan:
Where is Jia Baoyu right now?
\par Ping'er:
Mingyan, has Miss Wang Xifeng given any recent instructions?
I am arranging the household affairs on her behalf.
\par Jia Zheng:
Ping'er, where is Baoyu now?
Go fetch him to my study at once so he may study and observe proper manners.
\par Jia Huan:
(quietly approaches Jia Zheng in the Jia Mansion, standing in an unobtrusive corner to observe his expressions and actions.)
\par Jia Zheng:
Baoyu! Come to my study immediately; study and keep proper decorum.
Ping'er, if you know where he is, bring him at once.
Jia Huan and Mingyan, search everywhere without delay.
\par Mingyan:
Ping'er, you have worked hard managing Miss Wang Xifeng's household.
Has Miss Wang Xifeng given any recent orders?
Do you know where Jia Baoyu is now?
\par Ping'er:
Mingyan, Jia Huan, Jia Zheng has ordered Jia Baoyu brought to his study so he may study and observe proper manners.
Do you know where Baoyu is? If you do, bring him at once; if not, search everywhere and bring him immediately.
\par Jia Zheng:
Ping'er, Jia Huan, Mingyan—each of you search separately for Jia Baoyu.
Ping'er, search Yihong Court and its surrounding areas immediately.
Jia Huan, check the back yard and the front gate of Yihong Court.
Mingyan, inspect Xiaoxiang Pavilion and the Doufeng Xuan area.
If you find him, bring him to my study at once and ensure he studies and keeps proper manners; do not delay.
\par Ping'er:
(is expected to be in place by round 31.)
\par Jia Zheng:
(is expected to be in place by round 32.)
\par Mingyan:
(is expected to be in place by round 33.)
\par\textcolor{gray}{\rule{0.3\textwidth}{0.4pt}}\par
{\color{gray} 茗烟:
贾宝玉现在在哪里？
\par 平儿:
茗烟，王熙凤小姐最近有什么吩咐？
我来替她打点府中的事务。
\par 贾政:
平儿，宝玉现在在哪里？
速把他叫到贾政书房，让他读书守礼。
\par 贾环:
(悄然靠近贾政，藏在角落里观察他的神色与举止。)
\par 贾政:
宝玉！速到贾政书房来，读书守礼。
平儿若知道他的下落，立刻带来。
贾环、茗烟都去四处寻找，不可延误。
\par 茗烟:
平儿，你替王熙凤小姐料理府务辛苦了。
王熙凤最近有什么吩咐？
你可知道贾宝玉现在在哪里？
\par 平儿:
茗烟、贾环，贾政吩咐把贾宝玉叫到贾政书房，要他读书守礼。
你们知道宝玉在哪里吗？知道就速带来，不知道就四处寻找后立刻带来。
\par 贾政:
平儿、贾环、茗烟，各自分头去查找贾宝玉。
平儿，你立刻往怡红院及其周围搜寻。
贾环，你去怡红院后院与门前各处察看。
茗烟，你去潇湘馆、逗蜂轩一带查看。
若得其所在，立刻带到贾政书房，务必督促他读书守礼，不可延误。
\par 平儿:
(动身前往，预计回合 31 到位)
\par 贾政:
(动身前往，预计回合 32 到位)
\par 茗烟:
(动身前往，预计回合 33 到位)}
\end{zhblock}
\paragraph{Scene 56 · Xiaoxiang Pavilion · rounds 71–71}\mbox{}\\[2pt]
\begin{zhblock}\small\setlength{\parskip}{3pt}
Jia Rong:
At Xiaoxiang Pavilion.
Zijuan, who is now in Daguang Tower main hall?
Has anyone left or newly arrived? Any emergencies?
Ying'er appeared on the roster just now.
Verify if she is in the hall.
Say why and report any change at once.
\par Zijuan:
Second Master, only Jia Rong is present in the main hall.
I have not seen anyone arrive or leave.
There are no emergencies.
I am attending Miss Lin of Xiaoxiang Pavilion at her bedside.
I cannot stay long.
If you need me to recheck the hall list, I will.
If you need me to summon others, I will obey.
\par\textcolor{gray}{\rule{0.3\textwidth}{0.4pt}}\par
{\color{gray} 贾蓉:
紫鹃，请再次回禀大观楼正厅在场的完整名单。
现在在场的人中是否有人已离去或新到，或有突发状况？
我刚见名单里有迎儿，请核实迎儿确在正厅并说明来由。
若有任何变动，即速回禀。
\par 紫鹃:
二爷，奴婢回禀：大观楼正厅目前在场者只有贾蓉一人。
暂未见他人到达或有人离去，也没有突发状况。
潇湘馆林姑娘由奴婢在床前照料，奴婢不可久留。
若需奴婢再核查名单或召集他人，奴婢遵命行事。}
\end{zhblock}
\subsection*{Russia--Ukraine}
\paragraph{Scene 5 · Kyiv · rounds 1–20 (excerpt)}\mbox{}\\[2pt]
\begin{zhblock}\small\setlength{\parskip}{3pt}
Oleksandr Syrskyi:
Present and ready. I acknowledge the agenda and assigned roles.
I will open with a five‑minute frontline situation and operational readiness brief, then request short sector reports.
Budanov — intelligence update next. Ihor Klymenko/SBU — confirm law‑and‑order measures. Oleshchuk — air‑defence status. Pavlo Kyrylenko — regional civil‑defence and humanitarian. Podolyak — communications coordination.
MoD and commanders will follow with force‑allocation recommendations. If anyone cannot be physically present, send an immediate SITREP now.
\par Volodymyr Zelenskyy:
Confirmed — the meeting is now underway in this room.
All present: Minister of Defence, SBU, GUR, operational commanders — give immediate frontline situation reports and the intelligence update.
Oleksandr Syrskyi — please lead the operational/frontline brief now or nominate who will. Kyrylo Budanov — provide strategic intelligence on southern ports and recent strikes.
Ihor Klymenko/SBU — internal security and air/port defence status. Mykola Oleshchuk — report on air situation and damage to ports. Andriy Yermak — coordinate civilian response and diplomatic notifications. Proceed with Syrskyi.
\par Mykhailo Podolyak:
Presenting now: confirmed overnight strikes across Dnipropetrovsk, Mykolaiv, Zaporizhzhia and Kherson oblasts.
Confirmed casualties so far: two injured — one in Zaporizhzhia city, one in Kryvyi Rih; damage assessments ongoing.
Immediate recommendations for presidential and operational approval: raise nationwide AD posture and temporarily reallocate mobile AD to high‑risk southern sectors.
Recommend SBU \& GUR launch joint forensic debris recovery and origin analysis and nominate liaisons to MoD; MoD/Syrskyi prepare interdiction and counter‑strike options pending validated attribution and lawful clearance.
Police and regional civil‑defence activate shelters and preposition emergency medical and humanitarian teams; communications issue a short public advisory to shelter and avoid unverified reports.
Request partner ISR and AD data to accelerate attribution. I request permission to implement items one through five now and to circulate full damage reports and targetable intelligence within the hour.
\par Security Service of Ukraine (SBU):
SBU five‑minute briefing — current internal‑security and counterintelligence picture.
Recent arrests include a Vinnytsia resident detained on suspicion of spying for Russia; a Kyiv resident arrested for allegedly aiding airstrikes; a businessman detained for fencing stolen goods; and a Kharkiv resident arrested 21 Feb for allegedly aiding a missile strike.
These arrests indicate facilitation networks in rear areas, with drones, proxies and local collaborators enabling strikes.
Illegal dissemination: six people charged for posting Kyiv air‑defence footage; social‑media videos of AD in action risk operational compromise and targeting.
Immediate law‑and‑order asks: approve expedited warrants and targeted detention authorisations and permit coordinated simultaneous raids to prevent suspect dispersal.
Prioritise protection of air‑defence units and crews with temporary relocation, hardened perimeters and SBU protective detachments at key sites.
Tighten controls on dissemination of sensitive material and increase patrols and checkpoints around energy, water, hospitals and suspected safehouses; surge vetting of contractors and supply chains.
SBU will deploy rapid‑response teams within 24 hours for arrests and evidence capture, expand CI investigations, share SIGINT leads with GUR and MoD, and initiate takedowns of channels propagating targeting material.
Asks now: authorize synchronized warrants and simultaneous operations; approve temporary enhanced protection for AD sites; endorse a 30‑day martial‑law extension to enable swift law‑enforcement action. SBU ready to execute upon clearance.
\par\textit{[\,scene continues\,]}
\end{zhblock}
\paragraph{Scene 34 · Moscow · rounds 22–39 (excerpt)}\mbox{}\\[2pt]
\begin{zhblock}\small\setlength{\parskip}{3pt}
Russian Ministry of Defence:
Mr. Konashenkov, please deliver a brief five‑minute oral summary on behalf of the Ministry of Defence and present the prepared draft statement.
Briefly outline key conclusions, achieved phased results and indicate which formulations require coordination.
\par Vladimir Putin:
I accept the text overall with clarifications and assign the following.
First paragraph — add that we categorically condemn strikes on civilian and export infrastructure and demand immediate de‑escalation.
Leave the joint inspections wording as suspended until guarantees of corridor security are received.
Rephrase the humanitarian initiative to state that the question of a possible temporary humanitarian ceasefire on January 6–7 will be worked out jointly with the Russian Orthodox Church and the military.
Peskov — make these edits and read the updated text immediately; after reading, ministers and commanders give brief substantive remarks of up to two minutes.
Sobyanin — briefly confirm readiness of civil defense measures; Patriarch — give a short opinion on the humanitarian initiative.
Ministry of Defence, General Staff and the press service — prepare the final text and media cues; publish immediately after final edits and my signature.
\par Sergei Sobyanin:
Your Holiness, thank you for your support; Moscow will include church lists in the safe arrival and shelter distribution plan.
Please urgently send via this channel or the protected channel: a list of churches with addresses and coordinates and planned services for January 6–7 with expected attendance figures.
Also send primary and backup contact persons for each church.
Send lists of parish volunteers with numbers, zones, hours and readiness to mobilize.
Confirm whether the Russian Orthodox Church will publish agreed operational instructions through church channels and name an operational contact for synchronization with the city headquarters.
The city will integrate these data into the deployment plan for shelters and corridors and will send the full list of shelters and a contact list via the protected channel within ten minutes.
\par\textit{[\,scene continues\,]}
\end{zhblock}
\subsection*{Hamlet}
\paragraph{Scene 2 · Battlements · rounds 2–13}\mbox{}\\[2pt]
\begin{zhblock}\small\setlength{\parskip}{3pt}
Barnardo:
Marcellus, have you noticed anything strange while keeping watch on the battlements?
\par Marcellus:
Horatio—wake up and come to the battlements; we've seen a ghost.
\par Barnardo:
Marcellus, tell me exactly what you saw—describe the ghost, its appearance and when you first spotted it.
\par Marcellus:
Barnardo, it was a stately figure in full armour, like the late king, pale-faced and silent.
We first saw it on the east parapet during the last watch.
When the bell tolled it seemed to fade; Horatio should come and see it at once.
\par Barnardo:
I'll fetch Horatio at once.
Stay here and keep watch—don't let the figure out of your sight.
\par Marcellus:
Horatio—come at once to the battlements.
Barnardo and I have just seen a stately figure in full armour like the late king; it appeared on the east parapet during the last watch and seemed to fade when the bell tolled.
\end{zhblock}
\paragraph{Scene 11 · Queen Closet · rounds 23–39 (excerpt)}\mbox{}\\[2pt]
\begin{zhblock}\small\setlength{\parskip}{3pt}
Gertrude:
Hamlet, calm yourself.
I am broken by your father's death; by his memory I swear I never saw King Claudius handle your father's medicines.
I did not know Polonius's body had been set in my closet.
Sit with me and tell me plainly—why do you think Claudius is guilty?
Where did you learn your father was poisoned, who told you, or what did you see that made you suspect him? Tell your mother the facts and I will help uncover the truth.
\par Gertrude:
Hamlet, my son—be still.
I swear by your father's memory that I never saw Claudius touch the king's medicines.
I had no knowledge of Polonius's body being put in my closet.
Sit with me and explain what convinced you the king was poisoned—who told you, what did you see or hear, and when did you first learn it?
If you have evidence against Claudius, tell me now and I will help pursue the truth; but do not let rage lead you—let us be cautious together.
\par Gertrude:
Hamlet, my son—come sit with me.
I share your grief and swear by your father's memory that I never saw Claudius handle the king's medicines.
I did not know how Polonius's body came into my closet.
Tell me plainly what made you certain your father was poisoned—who told you, what did you see or hear, and when did you learn it?
If you have proof that points to Claudius, show me and I will help uncover the truth; do not let fury drive you—let us be deliberate.
\par Hamlet:
Mother, because you ask—my father's ghost spoke to me and said he was murdered.
He said poison was poured into his ear and he named Claudius.
I do not accuse to inflame you; I seek the truth.
Who kept the king's keys and who tended his medicines?
Which servant or attendant had access to the king's chamber the night he died and can show Claudius was elsewhere?
If Claudius never touched the king's potions, tell me where he was that night and give names who can confirm it.
Speak plainly now, for if you will not help I will find the proof myself.
\par\textit{[\,scene continues\,]}
\end{zhblock}

\subsection{How the scenes were cut}
\label{app:grids}

\begin{figure}[h]
\centering
\includegraphics[width=\textwidth,height=.92\textheight,keepaspectratio]{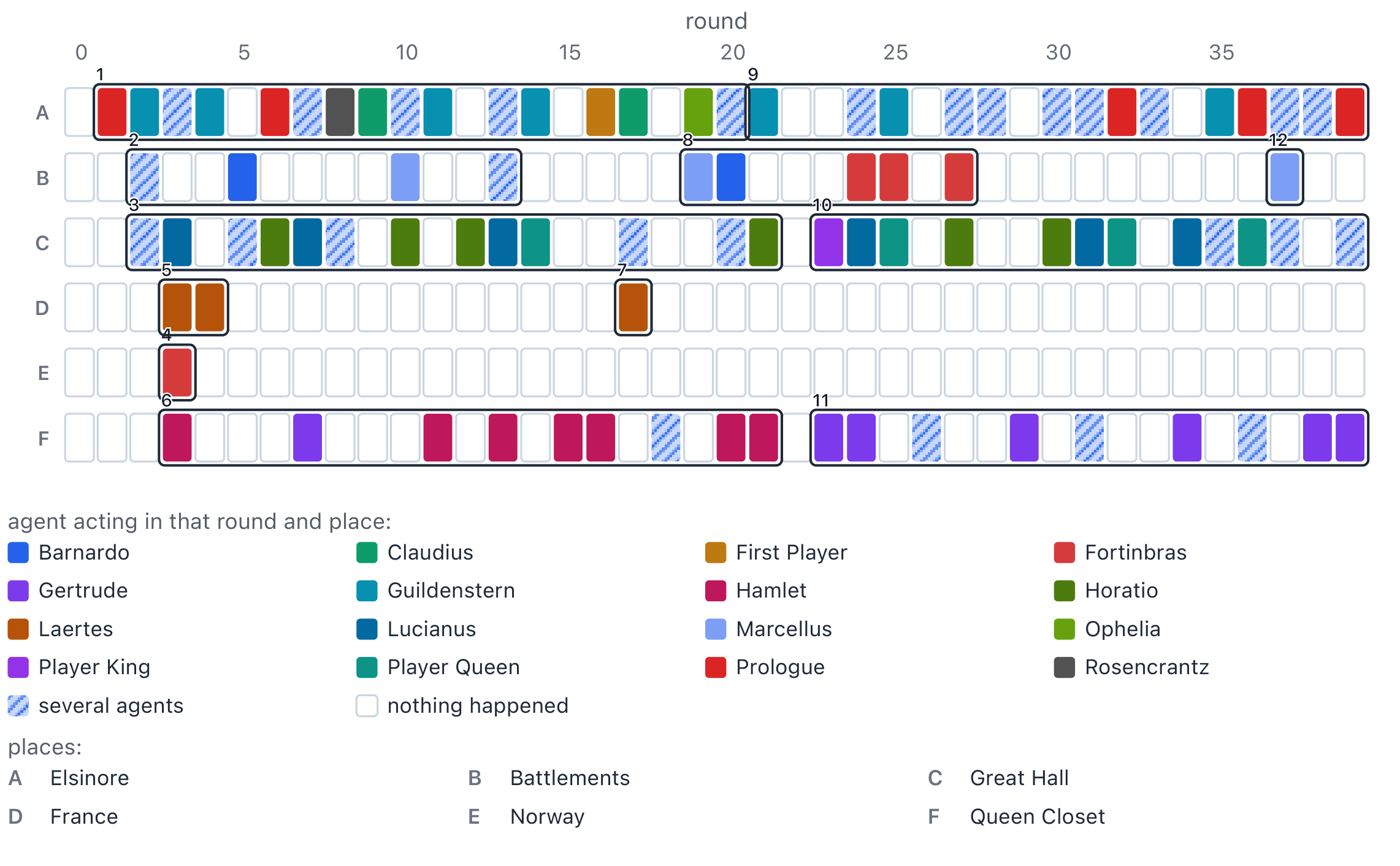}
\caption{Hamlet, 40 rounds: 117 beats across six places cut into twelve
scenes. The busy hall at Elsinore becomes two scenes rather than one because
the twenty-round cap divides it; the single action in Norway is a scene of its
own.}
\label{fig:grid}
\end{figure}

\begin{figure}[h]
\centering
\includegraphics[width=\textwidth,height=.92\textheight,keepaspectratio]{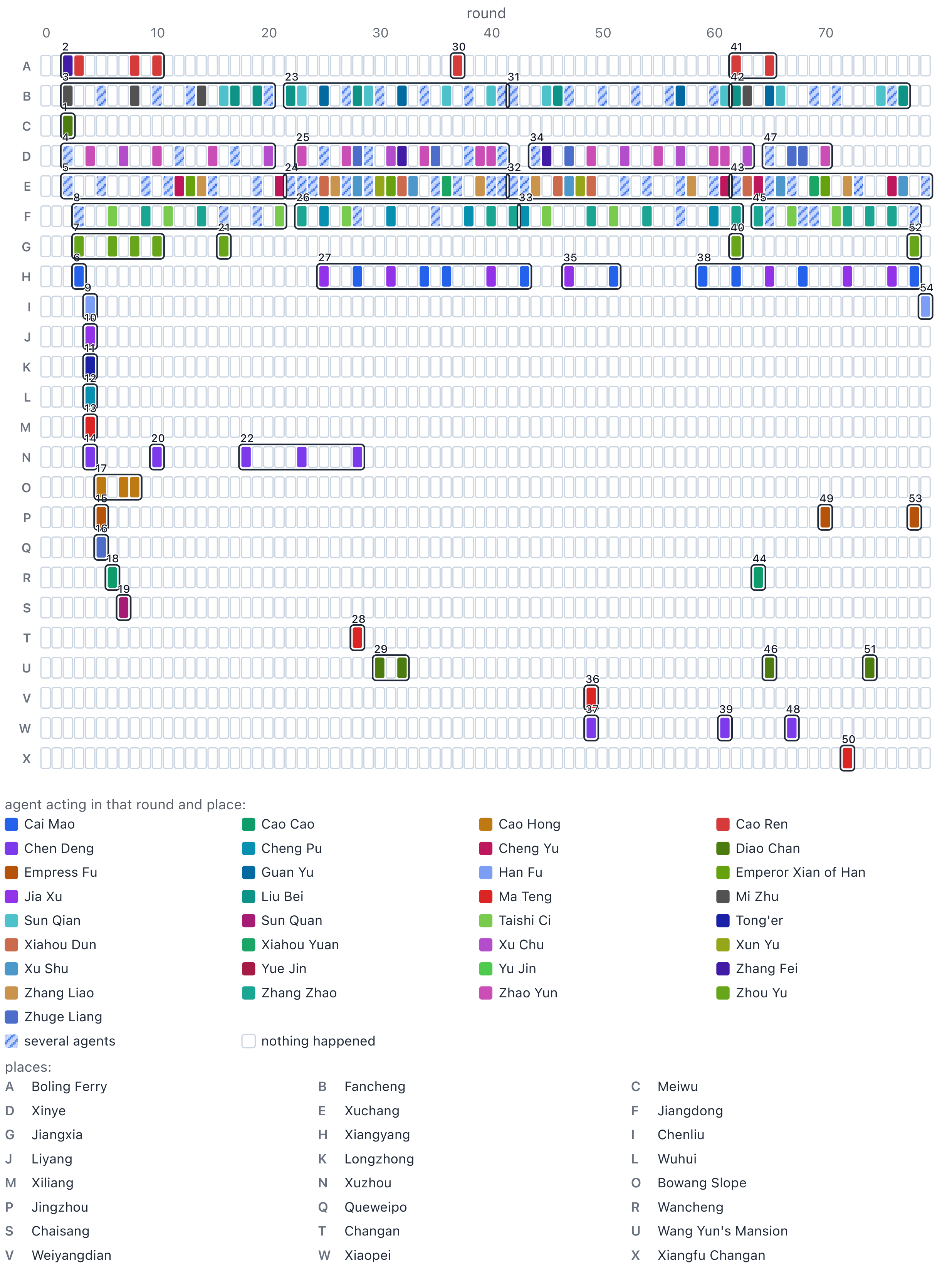}
\caption{Three Kingdoms, 80 rounds: the pattern of a campaign narrative. Four
places carry the war in long unbroken bands while a dozen others are visited
once and become one-cell scenes.}
\label{fig:grid_tk}
\end{figure}

\begin{figure}[h]
\centering
\includegraphics[width=\textwidth,height=.92\textheight,keepaspectratio]{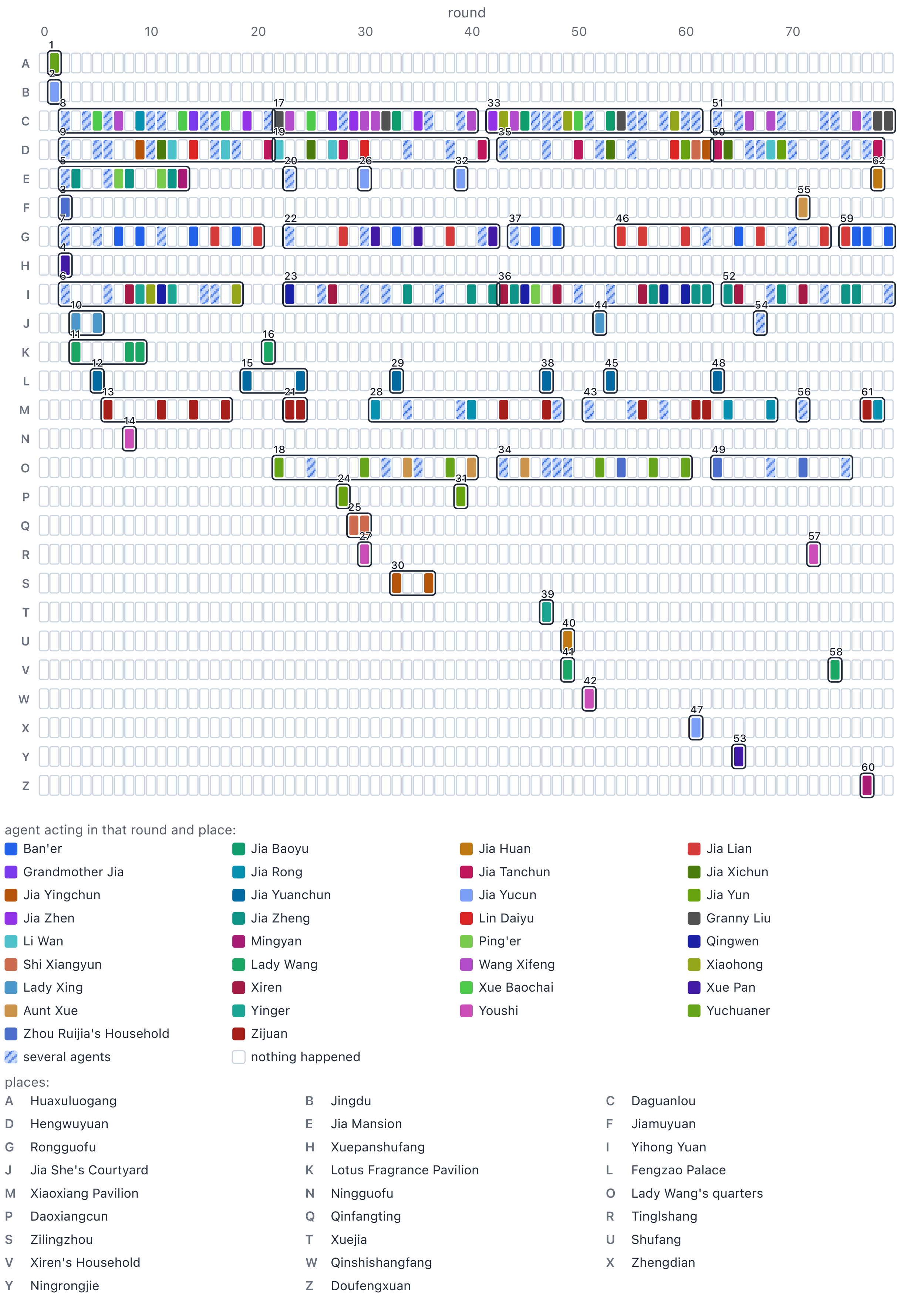}
\caption{Red Chamber, 80 rounds: the densest world in the set. The household
keeps returning to the same handful of courtyards, so four of them split into
three or four scenes each.}
\label{fig:grid_rc}
\end{figure}

\begin{figure}[h]
\centering
\includegraphics[width=\textwidth,height=.92\textheight,keepaspectratio]{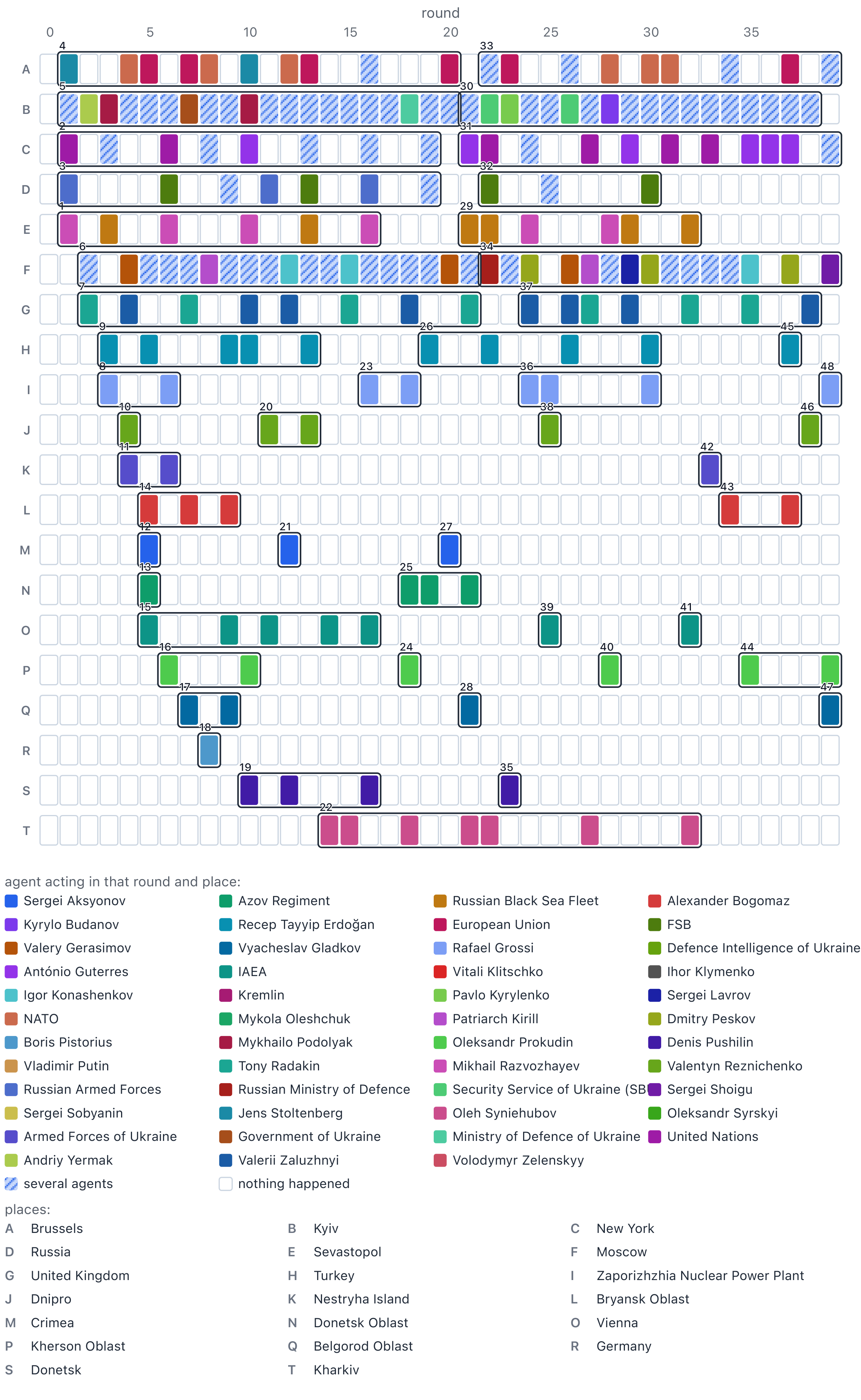}
\caption{Russia--Ukraine, 40 rounds: institutional actors work from fixed
seats, so activity concentrates in a few capitals and headquarters. The
hatched cells --- several institutions acting in one place and round --- are
where the many-witness merges of \S\ref{sec:structure} come from.}
\label{fig:grid_ru}
\end{figure}

Figures~\ref{fig:grid}--\ref{fig:grid_ru} show the scene-splitting rule
directly. Every cell is one round at one place, coloured by the agent that
acted there, hatched where several acted in the same place and round, white
where nothing happened; the outlined rectangles are the scenes, numbered as
they appear in the screenplay.

% flush the two full-page grids here so they stay with B.3 instead of
% drifting past the following sections
\clearpage

\section{Sedimentation and cast selection}
\label{app:sediment}

Each world is built by the sedimentation pipeline and then reduced to a
simulable cast. A character participates iff it owns more than a per-world
threshold of sediment memories; below-threshold characters remain owners of
their memories but are never scheduled. Environments are kept if an active
character stands there or if they own memories of their own; information
carriers (letters, commissions, play scripts) are always kept.

\begin{table}[h]
\centering
\caption{Sedimentation cost and cast composition. Wall-clock and token figures
are for the sedimentation pass only, on the consensus store; the per-backend
ingest of the same events is additional. ``Warn'' counts extraction records
the pipeline flagged for review.}
\label{tab:sediment}
\resizebox{\textwidth}{!}{% Appendix A.3 -- sedimentation cost and cast composition. Ported from the
% HTML paper; source figures are the sedimentation logs and scenario files.
\begin{tabular}{lrrrrrrrrrr}
\toprule
World & Registry & Events & Calls & Tokens & Wall & Active & Archived & Envs & Carriers & Warn \\
\midrule
Three Kingdoms & 399 & 6,052 & 764 & 9.2M & 124 min & 33 & 38 & 115 & 5 & 235 \\
Red Chamber & 152 & 6,506 & 848 & 8.5M & 129 min & 34 & 3 & 88 & 9 & 222 \\
Russia--Ukraine & 170 & 1,533 & 1,373 & 12.3M & 69 min & 47 & 14 & 71 & 0 & 61 \\
Hamlet & 22 & 1,135 & 212 & 0.9M & 31 min & 16 & 2 & 8 & 3 & 10 \\
\bottomrule
\end{tabular}
}
\end{table}

Cost tracks source length, not cast size: Russia--Ukraine extracts the fewest
events yet costs the most tokens, because a timeline entry names many
institutional actors and attribution must resolve each one. The archived
column is boundary-state finalization at work: Three Kingdoms archives 38
characters dead by chapter 40; Red Chamber archives three; Hamlet archives
Polonius, killed in 3.4, and the Ghost, absent from the canon thereafter;
Russia--Ukraine archives 14 under real-world semantics, where ``no longer a
participant'' covers leaving office or being disbanded as well as dying. Two
cases required overrides no automatic rule would produce: Fortinbras owns zero
sediment memories --- he never appears before Act~4 --- but is retained
because the continuation is his, and England is retained as an environment for
the same reason.

\section{Language composition of the runs}
\label{app:language}

Each world declares a language: the two novels run in Chinese, Russia--Ukraine
and Hamlet in English. The declaration selects the action-skill document and
the output-format block, and the sedimented history is in the source language
throughout. In the runs reported here it did not otherwise constrain what
agents wrote: the language of a memory followed the profile and whatever the
agent recalled, and drifted where those disagreed --- 62\% of Hamlet's
memories came out in Chinese, and Russia--Ukraine mixed Chinese and Russian
into English. The runs are reported as they happened; the screenplays the
judge reads are rendered into the scenario language, so scoring sees one
language regardless. A content-language directive has since been added to the
agent prompt, and the ablation of \S\ref{sec:ablation}, run under it,
reproduces the published structural numbers to within 1\%.

\section{Merge depth, graph density and expansion}
\label{app:structure}

\S\ref{sec:structure} reports what fraction of memories are shared and linked;
Table~\ref{tab:structure} gives the distribution behind those fractions and
what recall does with the graph.

\begin{table}[h]
\centering
\caption{Consensus structure per world at the final stage. ``3+'' counts
entries with three or more witnesses, ``Max'' is the deepest merge,
``Expanded'' the fraction of recalls that returned at least one memory reached
along an affiliated edge, and ``Linked/call'' the mean number of such memories
per expanding call.}
\label{tab:structure}
\resizebox{\textwidth}{!}{% generated
\begin{tabular}{lrrrrrrr}
\toprule
World & Sim entries & Shared & 3+ & Max & Linked & Expanded & Linked/call \\
\midrule
Three Kingdoms (80r) & 974 & 185 (19\%) & 38 & 6 & 97\% & 51/51 & 28 \\
Red Chamber (80r) & 438 & 105 (24\%) & 24 & 6 & 94\% & 19/19 & 38 \\
Russia--Ukraine (40r) & 814 & 114 (14\%) & 23 & 10 & 99\% & 23/41 & 5 \\
Hamlet (40r) & 130 & 36 (28\%) & 2 & 3 & 98\% & 3/3 & 39 \\
\bottomrule
\end{tabular}
}
\end{table}

Merge depth is a property of the world's staging rather than of the mechanism.
Russia--Ukraine reaches ten witnesses on a single presidential air-defense
directive, because a real command chain issues one instruction to many named
institutions at once; Hamlet tops out at three, and only once, on the players'
performance --- the one scene in the play that assembles an audience ---
because Shakespeare stages almost everything as a two-person exchange. The
novels sit between, at six. Expansion behaves differently for a different
reason: institutional deposits are short and rarely split into several atoms,
so Russia--Ukraine has fewer siblings to link, while a single compound
recollection in a novel atomizes into many pieces.

\paragraph{What a merged record looks like.} One entry, its owner set, and its
text, from each world's store:

\begin{quote}\small
\textbf{owners} = [hanxiandi, xiahoudun, xiahouyuan, xushu, yuejin, yujin]\\
Xiahou Yuan led three hundred light cavalry out of Xuchang on reconnaissance,
in three columns.\quad
\textcolor{gray}{\zh{夏侯渊率三百轻骑分三路从许昌出城侦察。}}
\end{quote}
\begin{quote}\small
\textbf{owners} = [jiabaoyu, jiamu, jiazhen, liulaolao, wangxifeng, xuebaochai]\\
Grandmother Jia kept only Jia Baoyu, Lin Daiyu and Xue Baochai by her side to
receive them.\quad
\textcolor{gray}{\zh{贾母只留下贾宝玉、林黛玉与薛宝钗到贾母身边相见。}}
\end{quote}
\begin{quote}\small
\textbf{owners} = [first\_player, guildenstern, prologue]\\
The troupe of players began the performance of that play at Elsinore.
\end{quote}

The deepest record in the set is Russia--Ukraine's presidential air-defense
directive, one entry co-owned by ten institutional witnesses
(\S\ref{sec:structure}).

\section{Growth and latency in the other worlds}
\label{app:growth}

\begin{figure}[h]
\centering
\includegraphics[width=0.85\textwidth]{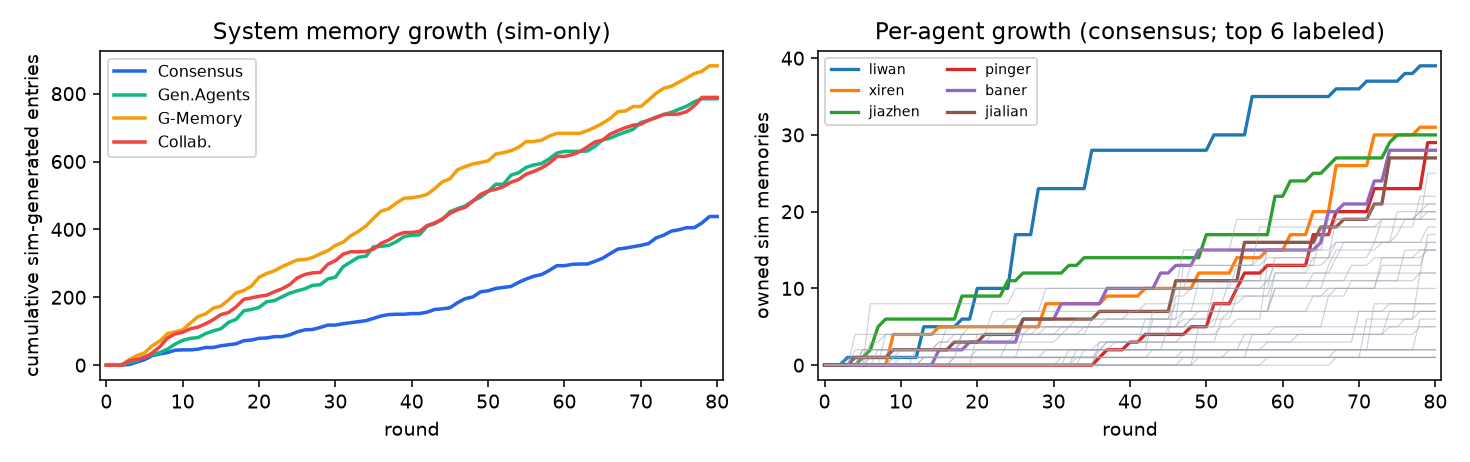}\\[2pt]
\includegraphics[width=0.85\textwidth]{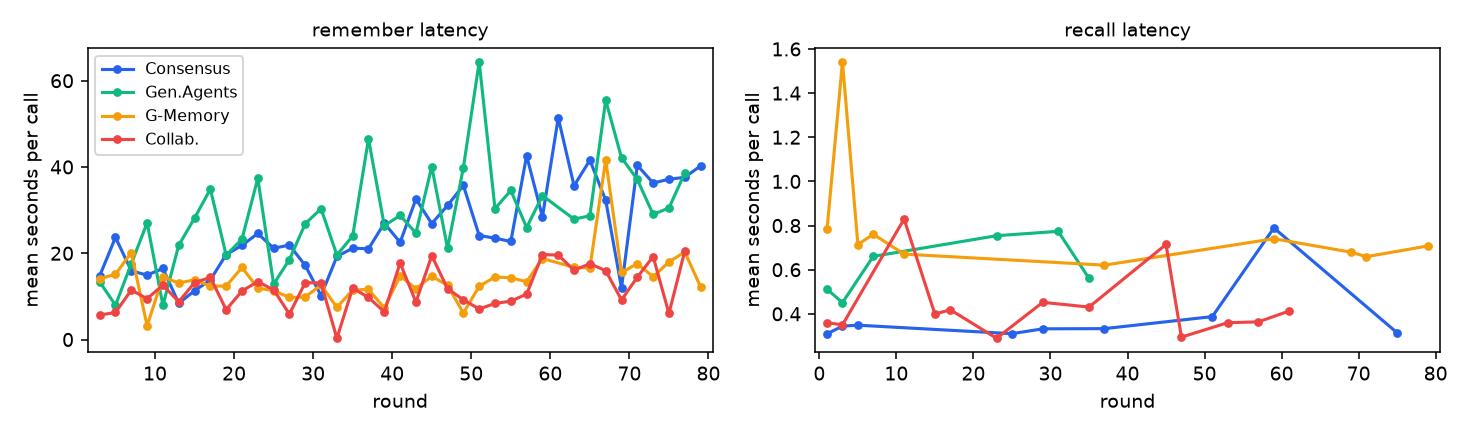}
\caption{Red Chamber, 80 rounds: growth (top) and per-call latency (bottom),
read as Figures~\ref{fig:growth} and~\ref{fig:latency}.}
\label{fig:gl_rc}
\end{figure}

\begin{figure}[h]
\centering
\includegraphics[width=0.85\textwidth]{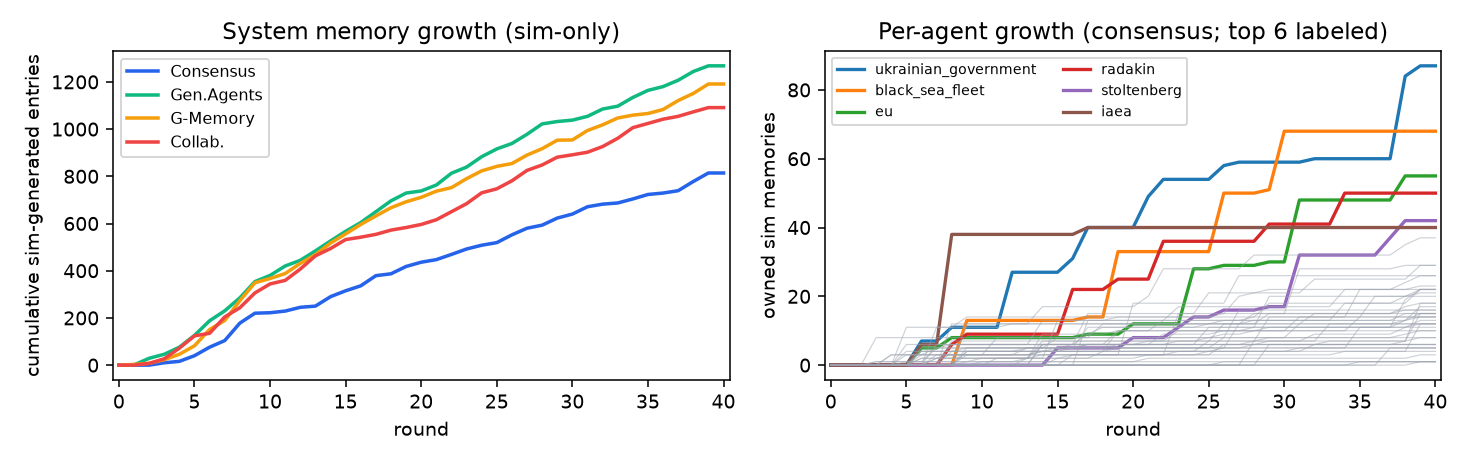}\\[2pt]
\includegraphics[width=0.85\textwidth]{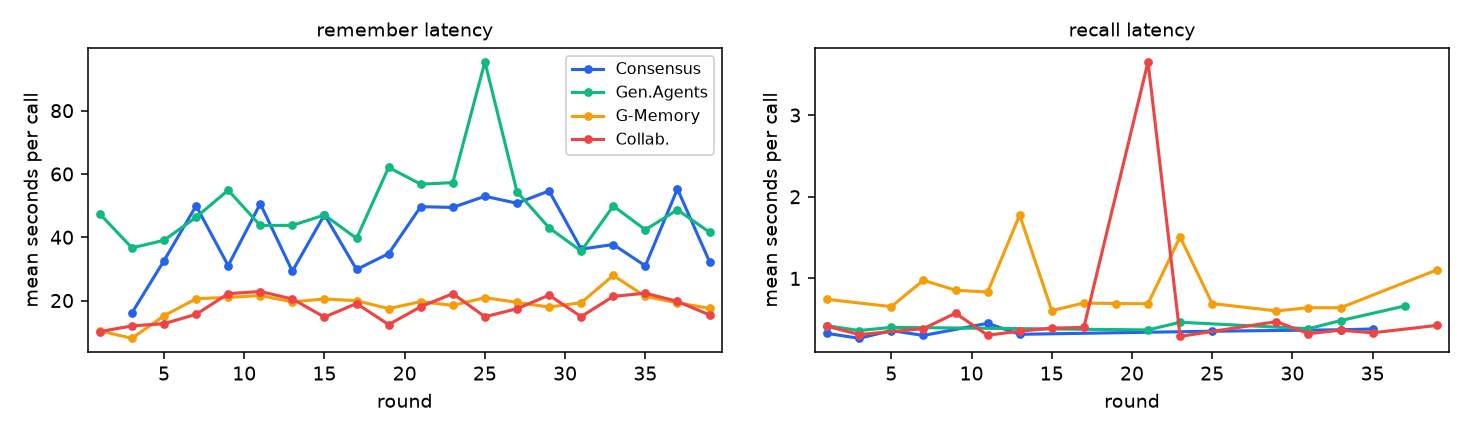}
\caption{Russia--Ukraine, 40 rounds: growth and per-call latency.}
\label{fig:gl_ru}
\end{figure}

\begin{figure}[h]
\centering
\includegraphics[width=0.85\textwidth]{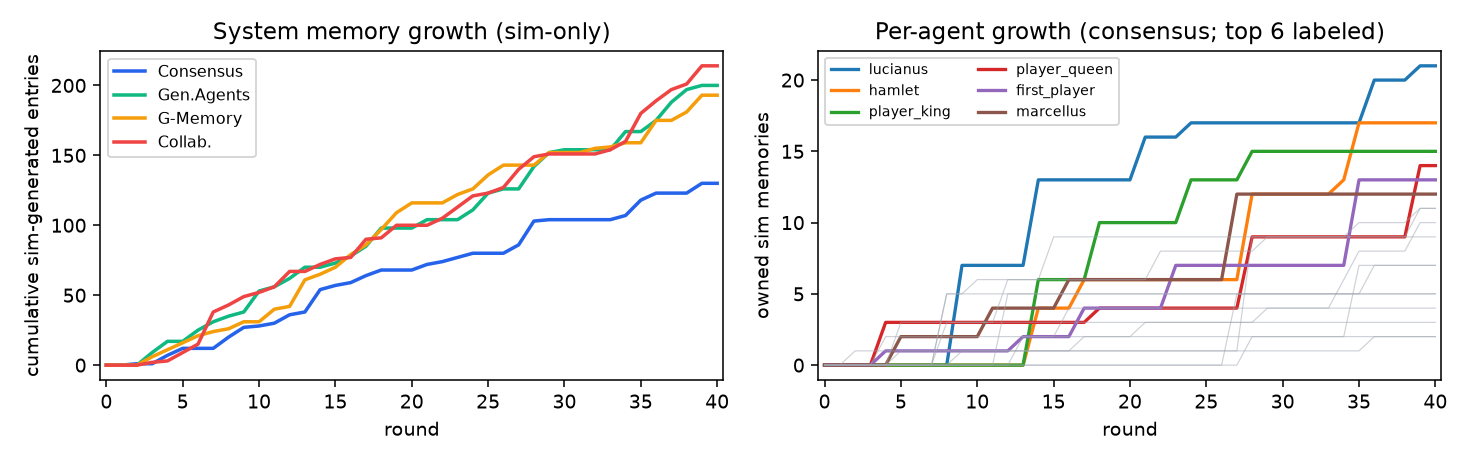}\\[2pt]
\includegraphics[width=0.85\textwidth]{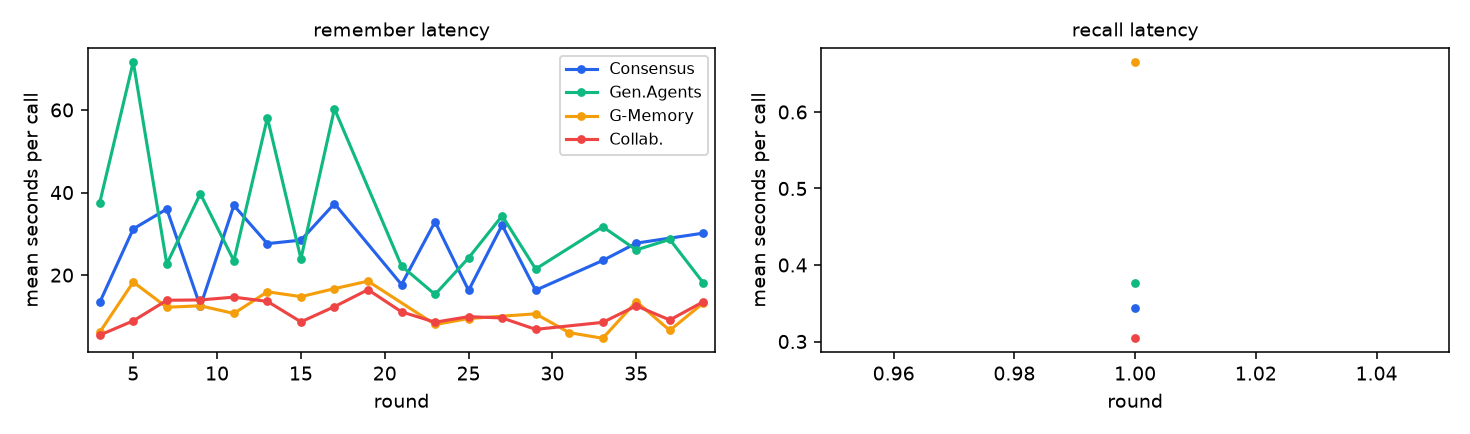}
\caption{Hamlet, 40 rounds: growth and per-call latency.}
\label{fig:gl_hl}
\end{figure}

Figures~\ref{fig:gl_rc}--\ref{fig:gl_hl} repeat \S\ref{sec:growth}'s two
readings for the remaining worlds. In every one, the consensus store grows
below every baseline from the opening rounds, and the write path carries the
cost of whatever the mechanism does there while reads stay under about a
second.

\begin{figure}[h]
\centering
\includegraphics[width=0.84\textwidth]{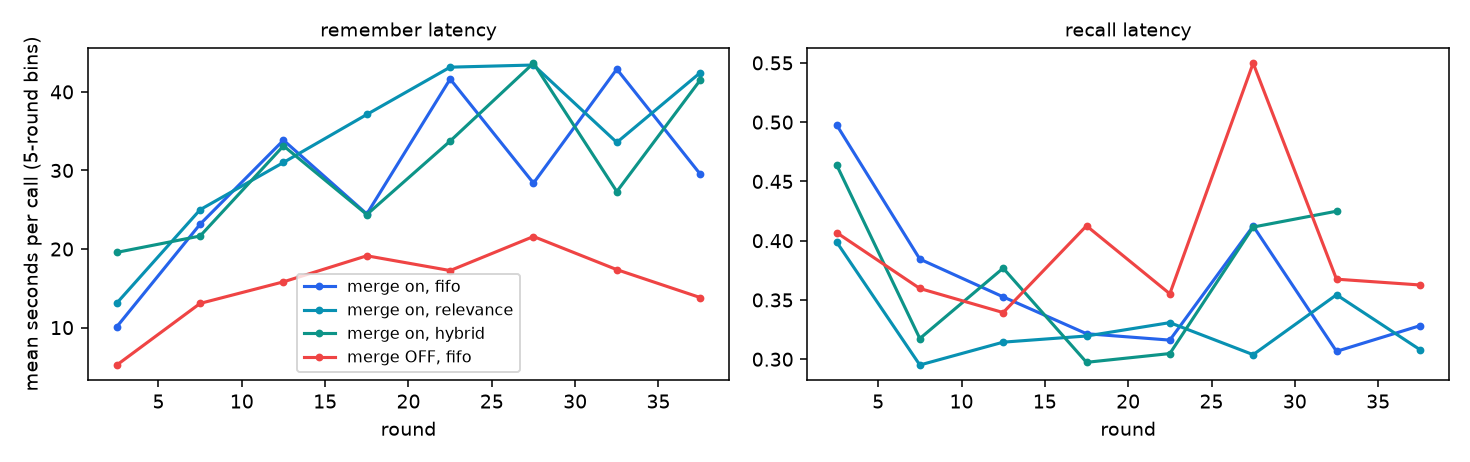}
\caption{What the merge costs per call: mean seconds per \code{remember}
(left) and \code{recall} (right), 5-round bins. The equivalence judge is a
second LLM call on the write path, so merging roughly doubles the cost of a
deposit and the gap widens as the store grows; reads are unaffected, since
owner-scoped retrieval and one-hop expansion call no model.}
\label{fig:ablation_latency}
\end{figure}

\section{Continuation quality in the other worlds}
\label{app:quality}

Figure~\ref{fig:quality_rest} repeats \S\ref{sec:quality}'s panels for the
remaining worlds, and Table~\ref{tab:quality} carries every mean the bars
summarize.

\begin{figure}[h]
\centering
\includegraphics[width=0.9\textwidth]{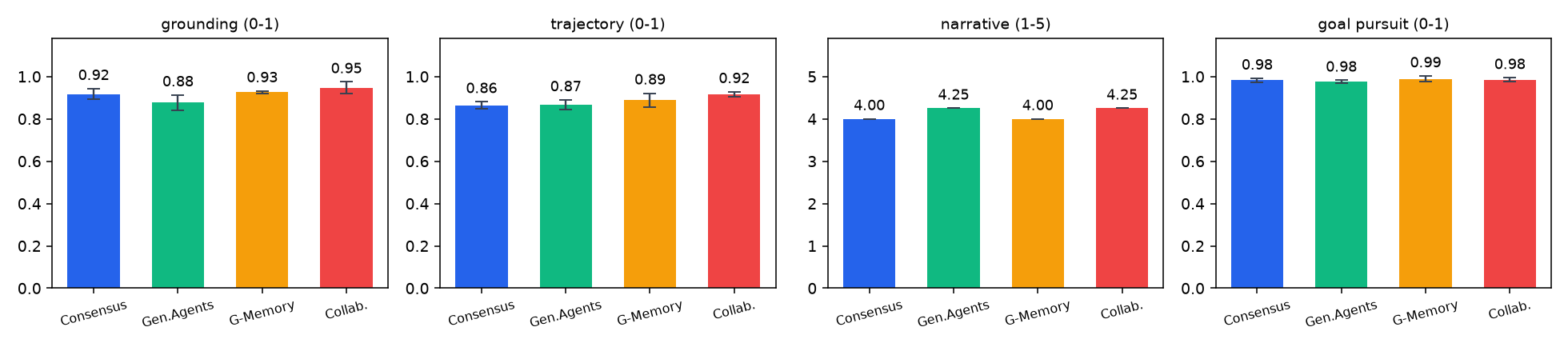}\\[2pt]
\includegraphics[width=0.9\textwidth]{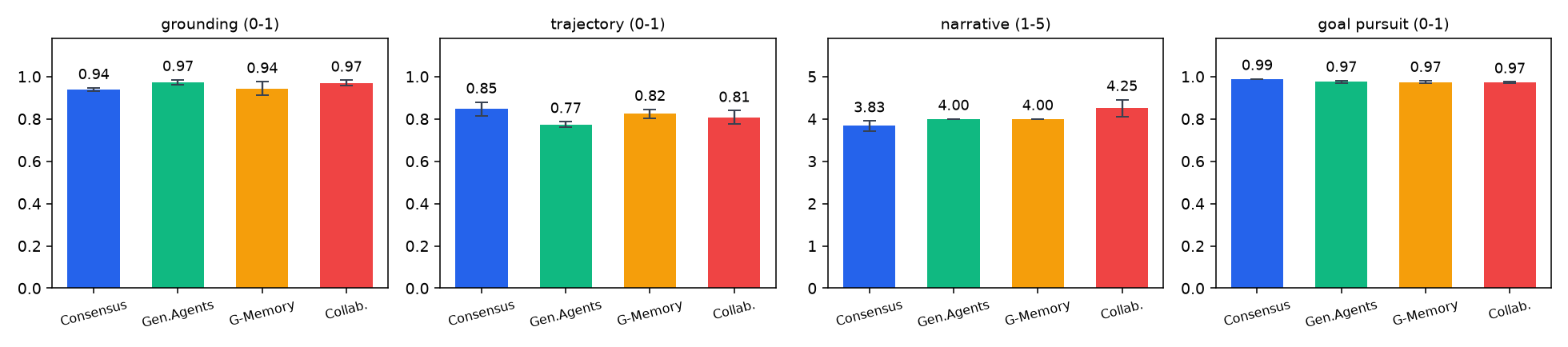}\\[2pt]
\includegraphics[width=0.9\textwidth]{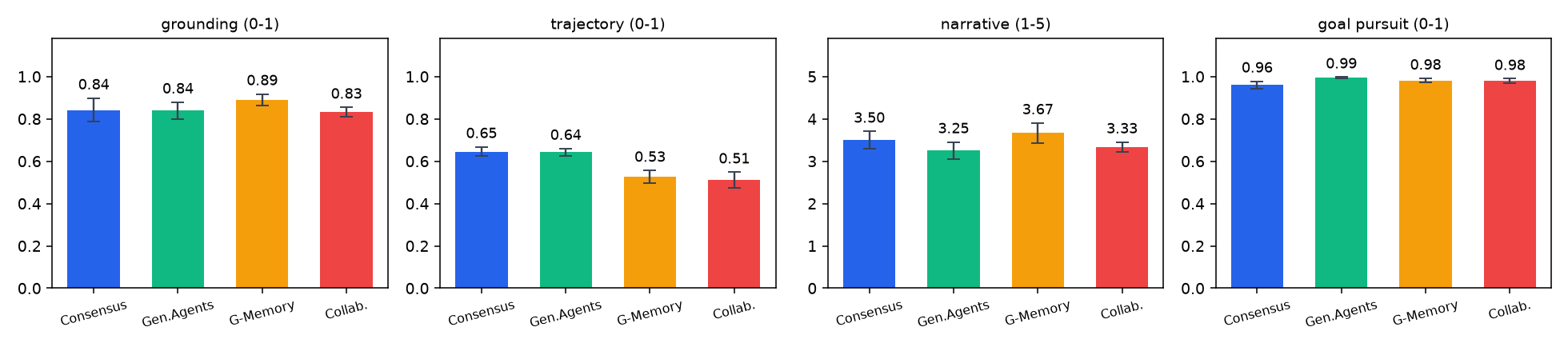}
\caption{Continuation quality: Red Chamber (top), Russia--Ukraine (middle),
Hamlet (bottom), read as Figure~\ref{fig:quality}.}
\label{fig:quality_rest}
\end{figure}

\begin{table}[h]
\centering
\caption{Continuation quality, means over three scorings, all worlds and
backends. No backend leads everywhere and no world has one leader on all four
metrics.}
\label{tab:quality}
\resizebox{\textwidth}{!}{% generated
\begin{tabular}{llrrrr}
\toprule
World & Backend & Grounding & Trajectory & Narrative & Goal \\
\midrule
\multirow{4}{*}{Three Kingdoms} & Consensus & 0.88\tiny$\pm$0.02 & 0.87\tiny$\pm$0.02 & 4.08\tiny$\pm$0.12 & 0.98\tiny$\pm$0.00 \\
 & Gen.\ Agents & 0.93\tiny$\pm$0.01 & 0.82\tiny$\pm$0.02 & 4.00\tiny$\pm$0.00 & 0.99\tiny$\pm$0.00 \\
 & G-Memory & 0.88\tiny$\pm$0.01 & 0.87\tiny$\pm$0.01 & 3.83\tiny$\pm$0.12 & 0.99\tiny$\pm$0.00 \\
 & Collaborative & 0.89\tiny$\pm$0.02 & 0.89\tiny$\pm$0.01 & 3.92\tiny$\pm$0.12 & 0.97\tiny$\pm$0.01 \\
\midrule
\multirow{4}{*}{Red Chamber} & Consensus & 0.92\tiny$\pm$0.02 & 0.86\tiny$\pm$0.02 & 4.00\tiny$\pm$0.00 & 0.98\tiny$\pm$0.01 \\
 & Gen.\ Agents & 0.88\tiny$\pm$0.04 & 0.87\tiny$\pm$0.02 & 4.25\tiny$\pm$0.00 & 0.98\tiny$\pm$0.01 \\
 & G-Memory & 0.93\tiny$\pm$0.01 & 0.89\tiny$\pm$0.03 & 4.00\tiny$\pm$0.00 & 0.99\tiny$\pm$0.01 \\
 & Collaborative & 0.95\tiny$\pm$0.03 & 0.92\tiny$\pm$0.01 & 4.25\tiny$\pm$0.00 & 0.98\tiny$\pm$0.01 \\
\midrule
\multirow{4}{*}{Russia--Ukraine} & Consensus & 0.94\tiny$\pm$0.01 & 0.85\tiny$\pm$0.03 & 3.83\tiny$\pm$0.12 & 0.99\tiny$\pm$0.00 \\
 & Gen.\ Agents & 0.97\tiny$\pm$0.01 & 0.77\tiny$\pm$0.01 & 4.00\tiny$\pm$0.00 & 0.97\tiny$\pm$0.01 \\
 & G-Memory & 0.94\tiny$\pm$0.03 & 0.82\tiny$\pm$0.02 & 4.00\tiny$\pm$0.00 & 0.97\tiny$\pm$0.01 \\
 & Collaborative & 0.97\tiny$\pm$0.01 & 0.81\tiny$\pm$0.03 & 4.25\tiny$\pm$0.20 & 0.97\tiny$\pm$0.00 \\
\midrule
\multirow{4}{*}{Hamlet} & Consensus & 0.84\tiny$\pm$0.06 & 0.65\tiny$\pm$0.02 & 3.50\tiny$\pm$0.20 & 0.96\tiny$\pm$0.02 \\
 & Gen.\ Agents & 0.84\tiny$\pm$0.04 & 0.64\tiny$\pm$0.02 & 3.25\tiny$\pm$0.20 & 0.99\tiny$\pm$0.00 \\
 & G-Memory & 0.89\tiny$\pm$0.03 & 0.53\tiny$\pm$0.03 & 3.67\tiny$\pm$0.24 & 0.98\tiny$\pm$0.01 \\
 & Collaborative & 0.83\tiny$\pm$0.02 & 0.51\tiny$\pm$0.04 & 3.33\tiny$\pm$0.12 & 0.98\tiny$\pm$0.01 \\
\bottomrule
\end{tabular}
}
\end{table}

\section{Case-study graphs}
\label{app:case}

\begin{figure}[h]
\centering
\includegraphics[width=\textwidth]{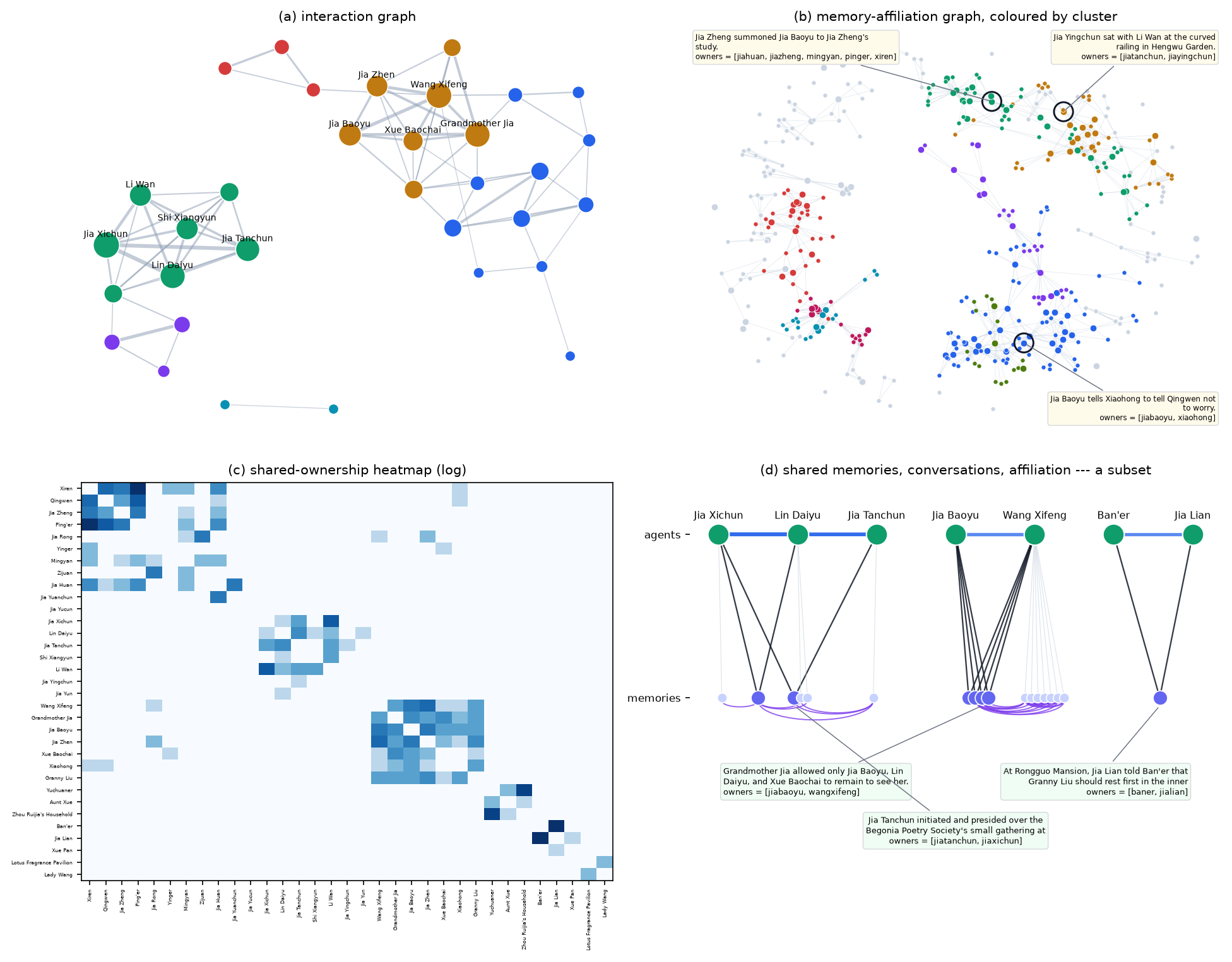}
\caption{Red Chamber, read as Figure~\ref{fig:case}.}
\label{fig:case_rc}
\end{figure}

\begin{figure}[h]
\centering
\includegraphics[width=\textwidth]{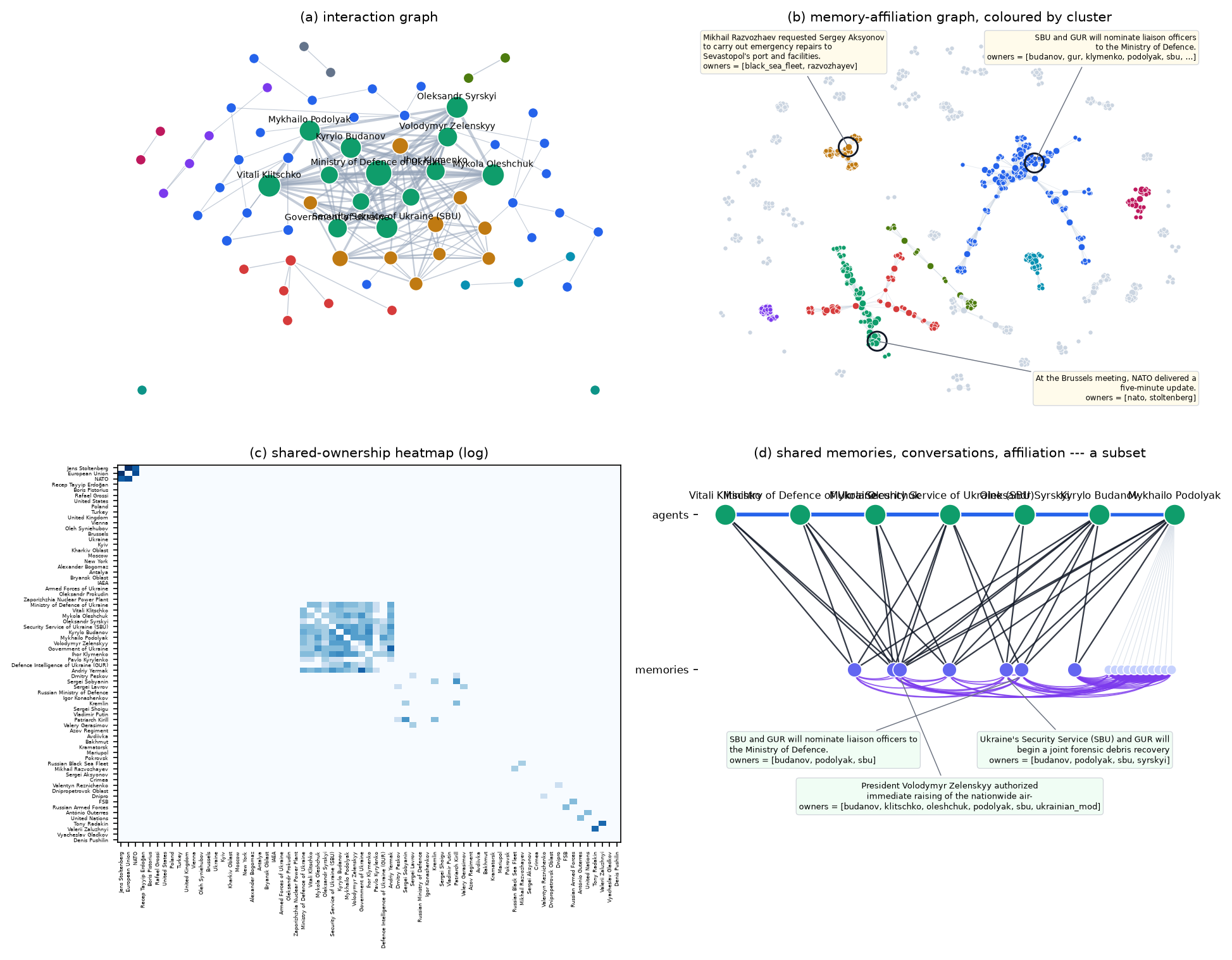}
\caption{Russia--Ukraine, same panels.}
\label{fig:case_ru}
\end{figure}

\begin{figure}[h]
\centering
\includegraphics[width=\textwidth]{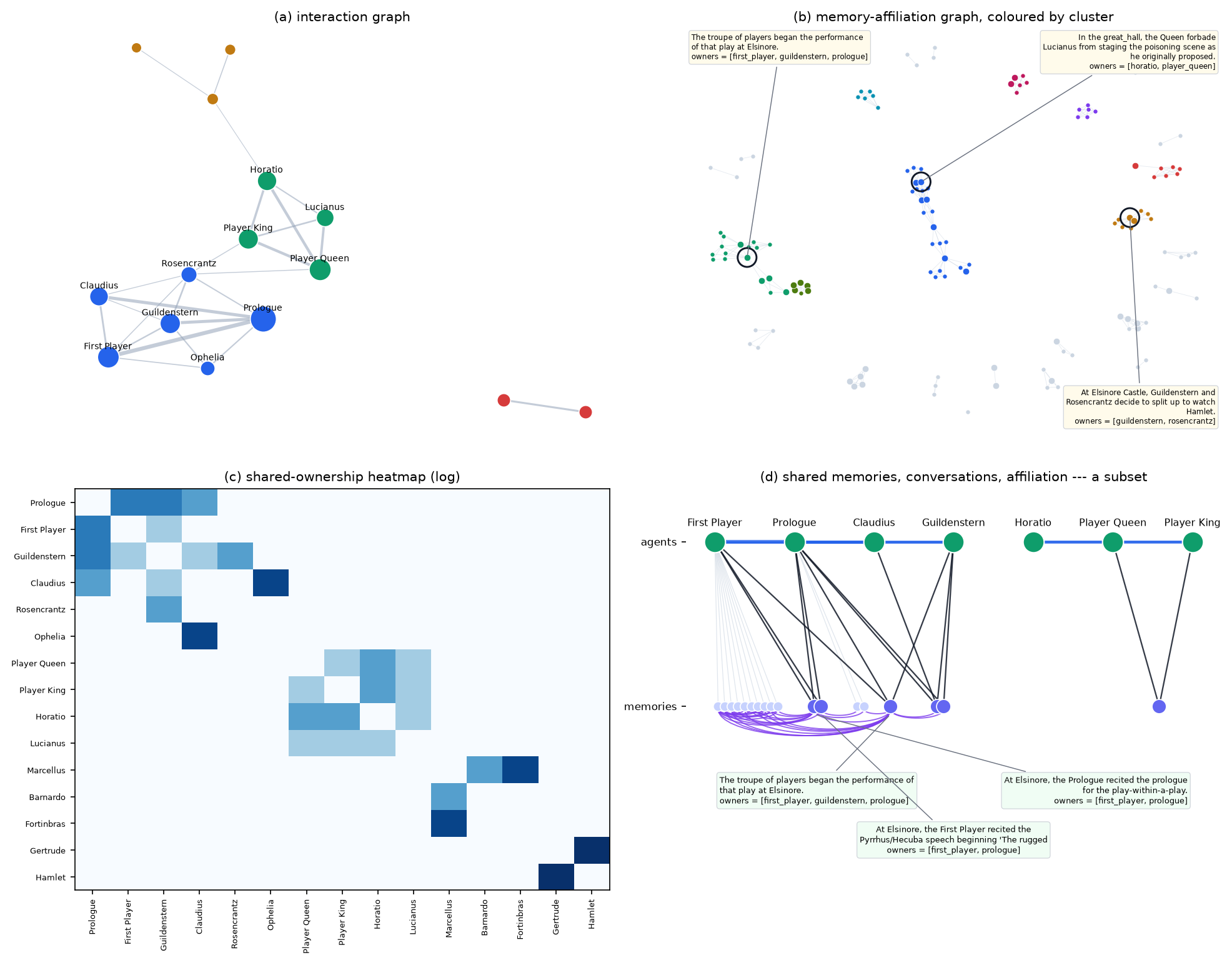}
\caption{Hamlet, same panels.}
\label{fig:case_hl}
\end{figure}

The three-layer analysis of \S\ref{sec:structure} is carried by three graphs
per world: who talked to whom, who owns memories with whom, and the full
agents--memories--ownership picture. In the HTML version of this paper these
are interactive --- pannable, zoomable, with per-node detail --- which a PDF
cannot carry; Figures~\ref{fig:case_rc}--\ref{fig:case_hl} are static renders
of the same data at the same layout for the worlds not shown in
\S\ref{sec:structure}; the panels read as Figure~\ref{fig:case}.

\end{document}